\documentclass[11pt]{article}

\usepackage[utf8]{inputenc}
\usepackage[T1]{fontenc}
\usepackage[margin=1in]{geometry}
\usepackage{amsmath,amssymb}
\usepackage{graphicx}
\usepackage{booktabs}
\usepackage{hyperref}
\usepackage{xcolor}
\usepackage{microtype}
\usepackage{float}

\usepackage{listings}
\usepackage{subcaption}

\usepackage[numbers,sort&compress]{natbib}

\newcommand{\benchmark}{Pencil Puzzle Bench}

\newcommand{\valFullPuzzles}{62,231}
\newcommand{\valFullVarieties}{94}
\newcommand{\valFullCreators}{814}
\newcommand{\valFullMedianMoves}{78}
\newcommand{\valFullMedianArea}{100}
\newcommand{\valGoldenPuzzles}{300}
\newcommand{\valGoldenVarieties}{20}
\newcommand{\valGoldenCreators}{118}
\newcommand{\valGoldenMedianMoves}{54}
\newcommand{\valGoldenMedianArea}{100}
\newcommand{\valGoldenThirtyPuzzles}{30}
\newcommand{\valGoldenThirtyVarieties}{4}
\newcommand{\valGoldenThirtyCreators}{21}
\newcommand{\valGoldenThirtyMedianMoves}{89}
\newcommand{\valGoldenSixtyPuzzles}{60}

\newcommand{\valPuzzlesPerType}{15}
\newcommand{\valPerTypePPChange}{6.7}
\newcommand{\valPuzzlesPerTier}{5}
\newcommand{\valExpandedPuzzlesPerType}{3}
\newcommand{\valPzprjsTypes}{100}

\newcommand{\valNumModels}{51}
\newcommand{\valNumProviders}{11}
\newcommand{\valOpenAIConfigs}{14}
\newcommand{\valAnthropicConfigs}{11}
\newcommand{\valGoogleConfigs}{7}
\newcommand{\valXAIConfigs}{3}

\newcommand{\valDeepSeekConfigs}{2}
\newcommand{\valQwenConfigs}{5}
\newcommand{\valMinimaxConfigs}{2}
\newcommand{\valMistralConfigs}{2}
\newcommand{\valMoonshotConfigs}{2}
\newcommand{\valZhipuConfigs}{2}
\newcommand{\valXiaomiConfigs}{1}

\newcommand{\valAgenticRuns}{1,602}
\newcommand{\valAgenticBaselineModels}{45}
\newcommand{\valAgenticExcludedModels}{3}
\newcommand{\valAgenticTotalModels}{48}
\newcommand{\valAgenticExpandedModels}{3}
\newcommand{\valAgenticMeanTurns}{57.2}
\newcommand{\valAgenticMedianTurns}{29}
\newcommand{\valAgenticPNinetyTurns}{113}
\newcommand{\valAgenticMaxTurns}{1,221}
\newcommand{\valAgenticMeanDuration}{40}
\newcommand{\valAgenticMedianDuration}{17}
\newcommand{\valAgenticPNinetyDuration}{108}
\newcommand{\valAgenticMaxDuration}{14.3}

\newcommand{\valGptXhighDirect}{27.0}
\newcommand{\valGptXhighAgentic}{56.0}

\newcommand{\valGptXhighMatchedN}{84}
\newcommand{\valGptXhighMatchedDirect}{20.2}
\newcommand{\valGptXhighMatchedAgentic}{56.0}
\newcommand{\valGptXhighMatchedDelta}{+35.7}
\newcommand{\valGptHighDirect}{20.7}
\newcommand{\valGptHighAgentic}{36.7}
\newcommand{\valClaudeOpusDirect}{0.3}
\newcommand{\valClaudeOpusAgentic}{30.0}

\newcommand{\valClaudeOpusMatchedN}{30}

\newcommand{\valClaudeOpusMatchedDelta}{+30.0}

\newcommand{\valClaudeOpusMaxAgentic}{23.3}

\newcommand{\valClaudeSonnetDirect}{0.0}
\newcommand{\valClaudeSonnetAgentic}{3.3}

\newcommand{\valClaudeOpusFourFiveDirect}{0.3}
\newcommand{\valClaudeOpusFourFiveAgentic}{3.3}
\newcommand{\valClaudeOpusFourFiveThinkingDirect}{6.0}
\newcommand{\valClaudeOpusFourFiveThinkingAgentic}{6.7}
\newcommand{\valClaudeOpusThinkingDirect}{27.3}
\newcommand{\valClaudeOpusThinkingAgentic}{33.3}
\newcommand{\valClaudeOpusThinkingMatchedN}{84}
\newcommand{\valClaudeOpusThinkingMatchedDirect}{28.6}
\newcommand{\valClaudeOpusThinkingMatchedAgentic}{33.3}
\newcommand{\valClaudeOpusThinkingMatchedDelta}{+4.8}

\newcommand{\valReasoningScaling}{81}
\newcommand{\valGptFiveZeroDirect}{6.0}
\newcommand{\valGptFiveOneDirect}{7.7}
\newcommand{\valGptFiveTwoDirect}{9.3}
\newcommand{\valSudokuBestRate}{33.3}
\newcommand{\valTopNModels}{15}

\newcommand{\valXhighErrorRate}{35}
\newcommand{\valGptXhighErrThirty}{27.8}
\newcommand{\valGptXhighErrSixty}{39.6}
\newcommand{\valClaudeOpusErrThirty}{50.0}
\newcommand{\valClaudeOpusErrSixty}{31.2}
\newcommand{\valErrorClusterPct}{100}
\newcommand{\valErrorClusterLoHr}{1}
\newcommand{\valErrorClusterHiHr}{4}
\newcommand{\valPeakErrorPct}{62}
\newcommand{\valPeakLoHr}{2}
\newcommand{\valPeakHiHr}{3}
\newcommand{\valTimeoutEstLo}{2.5}
\newcommand{\valTimeoutEstHi}{3}
\newcommand{\valXhighGainOverHigh}{6}

\newcommand{\valUnsolvedPostAgentic}{49}

\newcommand{\valNewlySolvedPhrase}{3 previously unsolved varieties}

\newcommand{\valMoveCountRsq}{0.092}
\newcommand{\valTypeEtaSq}{0.242}
\newcommand{\valTypeVsMovesFactor}{2.6}
\newcommand{\valCompressionRsq}{0.385}
\newcommand{\valCompressionVsMovesFactor}{4.2}
\newcommand{\valCompressionAreaRsq}{0.415}
\newcommand{\valFullModelRsq}{0.621}

\newcommand{\valTotalCost}{28,246}
\newcommand{\valTotalRuns}{17,032}
\newcommand{\valLowestCostPerSuccess}{0.01}

\newcommand{\valCostVariation}{66,822}

\newcommand{\nummodels}{\valNumModels}
\newcommand{\numpuzzles}{\valGoldenPuzzles}
\newcommand{\numpuzzletypes}{\valGoldenVarieties}

\title{Pencil Puzzle Bench: A Benchmark for Multi-Step Verifiable Reasoning}

\author{Justin Waugh \\ \small Approximate Labs \quad \texttt{justin@approximatelabs.com}}

\date{}

\begin{document}

\maketitle

\begin{abstract}
We introduce \benchmark{}, a framework for evaluating large language model reasoning through pencil puzzles, a family of constraint-satisfaction problems closely related to NP-complete problems, with deterministic, step-level verification. From a database of \valFullPuzzles{} puzzles across \valFullVarieties{} varieties with verified unique solutions, we select a benchmark of \numpuzzles{} puzzles spanning \numpuzzletypes{} varieties and evaluate \nummodels{} models from \valNumProviders{} providers in two modes: \textit{direct ask} (single-shot) and \textit{agentic} (multi-turn with iterative verification). A key differentiator of our benchmark is that \textbf{every intermediate board state can be checked against variety-specific constraints}, localizing errors to the exact rule violated, providing the infrastructure for dense, per-move reward signals for process supervision and reinforcement learning.

Our evaluation reveals two distinct axes of capability: (1) \textit{reasoning effort scaling}, where GPT-5.2 improves \valReasoningScaling{}$\times$ from no reasoning to maximum effort; and (2) \textit{agentic iteration}, where Claude Opus 4.6 rises from \valClaudeOpusDirect{}\% to \valClaudeOpusAgentic{}\% through iterative checking, while GPT-5.2@xhigh improves from \valGptXhighMatchedDirect{}\% to \valGptXhighMatchedAgentic{}\%. Agentic attempts span a median of \valAgenticMedianTurns{} turns over \valAgenticMedianDuration{} minutes, with the longest exceeding \valAgenticMaxTurns{} turns and \valAgenticMaxDuration{} hours---a demanding test of long-context utilization, not just reasoning.
\end{abstract}

\begin{figure}[ht!]
\centering
\begin{minipage}[c]{0.49\textwidth}
    \includegraphics[width=\textwidth]{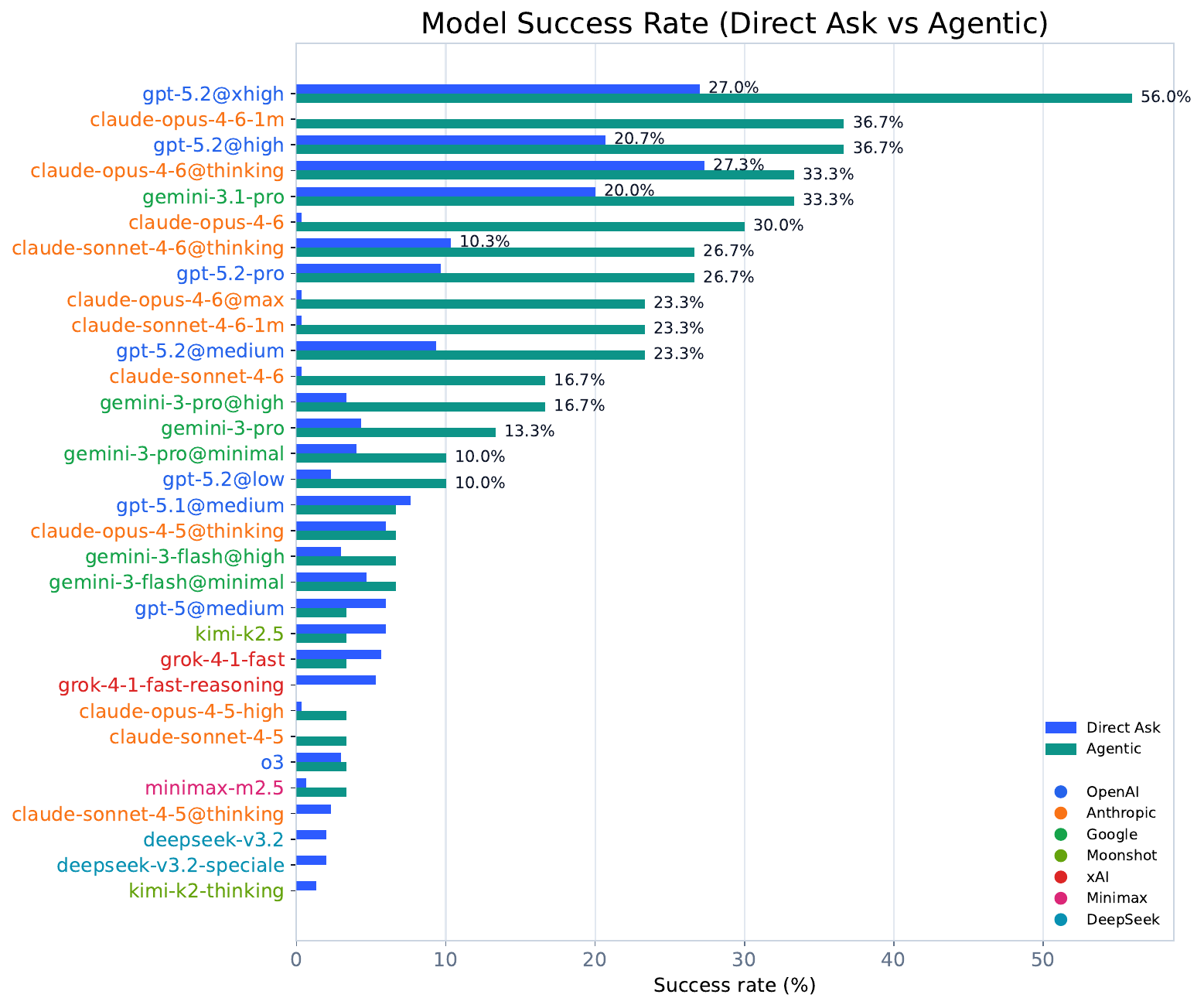}
\end{minipage}%
\hfill
\begin{minipage}[c]{0.47\textwidth}
    \centering
    \includegraphics[width=\textwidth,height=0.30\textheight,keepaspectratio]{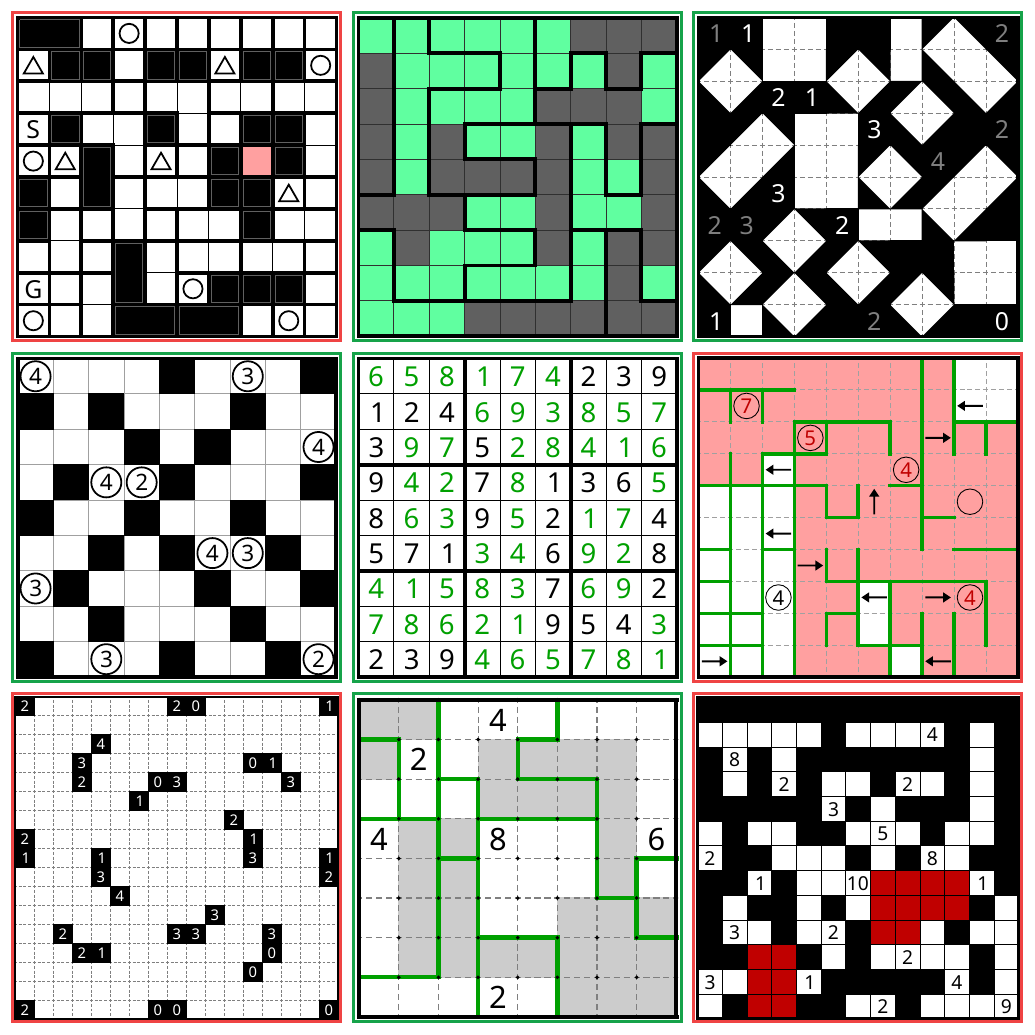}
\end{minipage}
\caption{Model success rates by strategy (left), with a 3$\times$3 gallery of partial and complete solves across diverse attempts (right).}
\label{fig:leaderboard}
\end{figure}

\section{Introduction}
\label{sec:intro}

The rapid progress of inference-time compute scaling, from chain-of-thought prompting~\cite{wei2022cot} to dedicated reasoning modes like OpenAI's o-series~\cite{openai2024o1} and DeepSeek-R1~\cite{deepseek2025r1}, has created urgent demand for benchmarks that can reliably measure multi-step reasoning. Existing benchmarks are either single-turn (GSM8K, MATH, ARC), increasingly contaminated~\cite{deng2023contamination,xu2024benchmarkleakage}, or cannot verify intermediate reasoning steps (SWE-bench checks final code, not reasoning). We need benchmarks that provide \textbf{multi-step, grounded, verifiable reasoning with step-level feedback}, enabling process reward models~\cite{lightman2023prm,zhang2025prm}, reinforcement learning from verifiable rewards (RLVR)~\cite{chen2025enigmata}, and curriculum learning.

Logic puzzles are an ideal testbed:
\begin{itemize}
    \item \textbf{Verifiable}: Solutions are unambiguously correct or incorrect; each puzzle has exactly one solution verified by a SAT solver
    \item \textbf{Multi-step}: Require chains of deductive reasoning across dozens to hundreds of moves
    \item \textbf{Step-level feedback}: Each move can be validated independently against variety-specific constraints, with error messages identifying \textit{which rule was broken and where}, e.g., ``Two shaded cells are adjacent'' in nurikabe or ``Loop crosses itself'' in slitherlink
    \item \textbf{Low contamination risk}: Puzzles are sourced from puzz.link, a Japanese puzzle community. While puzzle \textit{descriptions} may appear in training data as widely-shared URLs, \textit{solutions} are rarely published online, making solution contamination unlikely
    \item \textbf{Order-independent solutions}: The solution board state is unique, but the sequence of moves to reach it is flexible, allowing different orderings, mouse vs keyboard input, and undo/redo without affecting correctness
    \item \textbf{Rich representations}: Boards have ASCII text serialization (what models see), SVG vector rendering, and pixel-rasterized images, enabling future multimodal evaluation
\end{itemize}

Our contributions:
\begin{enumerate}
    \item A dataset of \valFullPuzzles{} puzzles across \valFullVarieties{} varieties with step-level solution traces, and an evaluation benchmark of \numpuzzles{} puzzles across \numpuzzletypes{} varieties with programmatic step-level verification, implemented as the \texttt{pencil-puzzle-bench} Python package
    \item Evaluation of \nummodels{} models from \valNumProviders{} providers in two modes (direct ask and agentic), revealing two distinct axes of capability improvement
    \item Discovery of the \textit{agentic gap}: agentic iteration benefits models across the capability spectrum, from +\valClaudeOpusThinkingMatchedDelta{}pp for models with extended thinking to +\valClaudeOpusMatchedDelta{}pp for models without
    \item Analysis of reasoning effort scaling, cost-efficiency, and infrastructure limits in agentic evaluation (median \valAgenticMedianTurns{} turns, up to \valAgenticMaxTurns{} turns per attempt)
    \item A difficulty analysis showing that solution compressibility ($R^2_{\text{adj}} = \valCompressionRsq{}$) predicts puzzle difficulty \valCompressionVsMovesFactor{}$\times$ better than move count ($R^2_{\text{adj}} = \valMoveCountRsq{}$)
\end{enumerate}

\section{Related Work}
\label{sec:related}

\paragraph{Reasoning Benchmarks}
GSM8K~\cite{cobbe2021gsm8k}, MATH~\cite{hendrycks2021math}, and ARC~\cite{clark2018arc} evaluate mathematical and scientific reasoning but are single-turn and have known contamination issues~\cite{deng2023contamination,xu2024benchmarkleakage,xu2024contaminationreview}. BIG-Bench~\cite{srivastava2022bigbench} and BBH~\cite{suzgun2022bbh} provide broader task coverage. RuleTaker~\cite{clark2020ruletaker} and StrategyQA~\cite{geva2021strategyqa} test logical and implicit reasoning. Recent work has raised concerns about benchmark integrity, evaluation awareness, and the limitations of static benchmarks~\cite{benchmarkinadequacies2024,statictodynamic2025,shi2025judgeagent}.

\paragraph{Inference-Time Compute and Reasoning Scaling}
The paradigm of scaling inference-time compute (rather than model size) has emerged as a dominant theme in 2025--2026. OpenAI's o-series models~\cite{openai2024o1} introduced adjustable reasoning effort, DeepSeek-R1~\cite{deepseek2025r1} demonstrated that RL can incentivize reasoning capabilities, and subsequent work has explored the frontiers of learning to reason and agentic systems~\cite{ke2025llmreasoning}. Complementary inference-time techniques improve reasoning without additional training, including self-consistency~\cite{wang2022selfconsistency}, least-to-most prompting~\cite{zhou2022leasttomost}, Plan-and-Solve~\cite{wang2023planandsolve}, Program of Thoughts~\cite{chen2022pot}, and deliberative search strategies such as Tree of Thoughts~\cite{yao2024tot} and Graph of Thoughts~\cite{besta2023got} (often paired with self-refinement~\cite{madaan2023selfrefine}). Process reward models~\cite{lightman2023prm,zhang2025prm,zheng2025prmsurvey} enable step-level verification for training. Our benchmark directly measures this trend: GPT-5.2's \valReasoningScaling{}$\times$ improvement across reasoning effort levels quantifies the returns to inference-time compute on a challenging domain.

\paragraph{Puzzle Benchmarks}
GridPuzzle~\cite{tyagi2024gridpuzzle} and ZebraLogic~\cite{lin2025zebralogic} evaluate logic puzzle solving but focus on text-based grid puzzles without step-level verification. Li et al.~\cite{li2024minesweeper} study Minesweeper as a reasoning testbed. PuzzleBench~\cite{zhang2025puzzlebench} provides dynamic multimodal puzzle evaluation. Recent puzzle-centric benchmarks emphasize synthetic verifiability and reflective solving (Enigmata~\cite{chen2025enigmata}, FINEREASON~\cite{chen2025finereason}) and challenge models with long-tail logic games (HardcoreLogic~\cite{liang2025hardcorelogic}); solver-in-the-loop work targets logic puzzle solving in declarative formalisms such as ASP~\cite{schrader2025solverloop}. Giadikiaroglou et al.~\cite{giadikiaroglou2024puzzlesurvey} survey puzzle-solving with LLMs. SATBench~\cite{wei2025satbench} generates puzzles from SAT formulas, while RiddleBench~\cite{halder2025riddlebench} tests generative reasoning. Most closely related, Sudoku-Bench~\cite{seely2025sudokubench} evaluates LLMs on 100 Sudoku variant puzzles sourced from Nikoli and competitive puzzle communities, testing ``eureka''-moment creative reasoning. Our work is distinguished by breadth (\valFullVarieties{} puzzle varieties in the full dataset, \numpuzzletypes{} evaluated, vs.\ Sudoku variants only), coordinate-based moves with step-level verification against variety-specific constraint sets, and the scale of our agentic evaluation (up to \valAgenticMaxTurns{} turns per attempt).

\paragraph{Tool Use, Agents, and Agentic Benchmarks}
Early work established the foundations: ReAct~\cite{yao2023react} combines reasoning with actions, Toolformer~\cite{schick2023toolformer} teaches models to use tools autonomously, and generative agents~\cite{park2023generativeagents} demonstrate complex agent behaviors. The field has since produced dedicated agentic benchmarks: SWE-bench~\cite{jimenez2024swebench} evaluates agents on real GitHub issues, tau-bench~\cite{yao2024taubench} tests multi-turn tool-agent-user interaction in real-world domains, TheAgentCompany~\cite{xu2024theagentcompany} evaluates agents on 175 professional tasks in a simulated company, and the Berkeley Function Calling Leaderboard~\cite{patil2025bfcl} benchmarks tool-calling capabilities. Recent surveys~\cite{guo2024multiagents,ke2025llmreasoning} categorize the rapidly growing space. We contribute a benchmark that evaluates models in both single-shot and agentic modes, with agentic solvers sustaining multi-turn interactions over dozens to hundreds of turns, testing long-context utilization and iterative reasoning on a domain with deterministic step-level verification.

\paragraph{Puzzle Complexity}
Many pencil puzzles are NP-complete~\cite{yato2003complexity,tang2022looppuzzles,nikolaev2025evolomino}, meaning that no efficient general-purpose algorithm is known. While individual instances may admit shortcuts, the theoretical richness of this domain makes it unlikely that surface-level pattern matching suffices across varieties and sizes.

\section{The \benchmark{} Framework}
\label{sec:benchmark}

\subsection{Technical Infrastructure}

\paragraph{The pzprjs Engine}
We build on pzprjs\footnote{\url{https://github.com/robx/pzprjs}}, a mature JavaScript framework for interactive pencil puzzles that powers the puzz.link community. Pzprjs implements \valPzprjsTypes{}+ puzzle varieties with full rule checking, error localization, and completion detection. Puzzles are shared as ``classic URLs,'' compact encodings widely used in the online puzzle community.

\paragraph{Solution Generation}
Solutions were generated using cspuz-solver2\footnote{\url{https://github.com/semiexp/cspuz-solver2}}, a SAT-based constraint solver that finds verified unique solutions. A key technical contribution is converting from the solver's output format (a complete board state) to the pzprjs move-set format (a sequence of interactive moves), bridging the constraint solver's representation to the puzzle engine's. This conversion enables fine-tuning on solution trajectories, bootstrapping capability in smaller models, analytics on puzzle difficulty, and verification that every puzzle has a unique valid solution. Step-level verification itself comes from the pzprjs \texttt{check()} function, which validates each intermediate board state against variety-specific constraints.

\paragraph{Board Representations}
Each puzzle board supports three representations:
\begin{enumerate}
    \item \textbf{ASCII text serialization}: The format models receive, a structured text encoding of cell contents, borders, and clues
    \item \textbf{SVG vector rendering}: A scalable vector graphic of the board state, suitable for display and debugging
    \item \textbf{Pixel images}: SVG can be rasterized to produce actual images, enabling future multimodal evaluation where models ``see'' the puzzle
\end{enumerate}

\subsection{Step-Level Verification}
\label{sec:verification}

A key contribution of \benchmark{} is \textbf{step-level verification}: the ability to validate not just final solutions, but every intermediate state during solving. Unlike benchmarks that only report ``correct'' or ``incorrect,'' our verifier identifies \textit{which specific constraint was violated and where}.

\paragraph{Variety-Specific Constraints}
Every variety in the dataset defines its own ordered constraint set. Rules are checked in order, with local/direct constraints (e.g., ``two shaded cells are adjacent'') evaluated before global state constraints (e.g., ``all cells must be filled,'' ``loop must be connected''). When only a global rule is broken, no intermediate move is known to violate a local sub-rule. When a direct constraint is broken, the engine provides immediate, localizable error feedback including error highlighting that identifies the offending cells. Examples:

\begin{itemize}
    \item \textbf{Nurikabe}: ``Shaded cells form a 2$\times$2 square'', ``Two numbered islands are connected'', ``Shaded region is not fully connected''
    \item \textbf{Slitherlink}: ``Loop branches at vertex'', ``Number clue unsatisfied (expected 2, got 3 edges)''
    \item \textbf{Norinori}: ``Shaded cell has no adjacent shaded cell'', ``Region contains fewer than 2 shaded cells''
    \item \textbf{Sudoku}: ``Duplicate number in row'', ``Duplicate number in block''
\end{itemize}

This per-constraint, per-cell error localization enables models (in agentic mode) to receive targeted feedback about \textit{what went wrong}, not just \textit{that something is wrong}. For RL training, each constraint violation provides the building blocks for dense, interpretable reward signals: the verification infrastructure exposes violation counts, types, and locations at each step, which can be composed into reward functions (e.g., violation-delta shaping, progress-based rewards).

\paragraph{Verification Pipeline}
Every move in \benchmark{} can be validated through a four-step pipeline:
\begin{enumerate}
    \item \textbf{Pre-move state}: Check current board for rule violations
    \item \textbf{Apply move}: Execute a coordinate-based action (e.g., \texttt{mouse,left,3,5})
    \item \textbf{Post-move validation}: Immediately check for new violations introduced by the move
    \item \textbf{Completion check}: Determine if the puzzle is fully and correctly solved
\end{enumerate}

\begin{figure}[ht!]
\centering
\begin{subfigure}[b]{0.3\textwidth}
    \includegraphics[width=\textwidth]{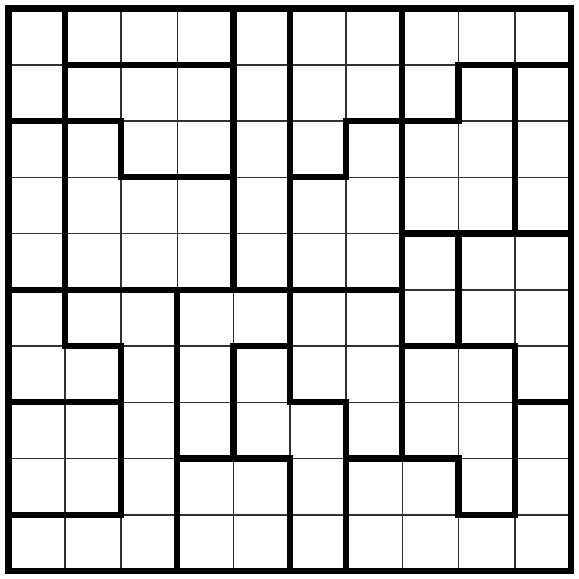}
    \caption{Initial state}
\end{subfigure}
\hfill
\begin{subfigure}[b]{0.3\textwidth}
    \includegraphics[width=\textwidth]{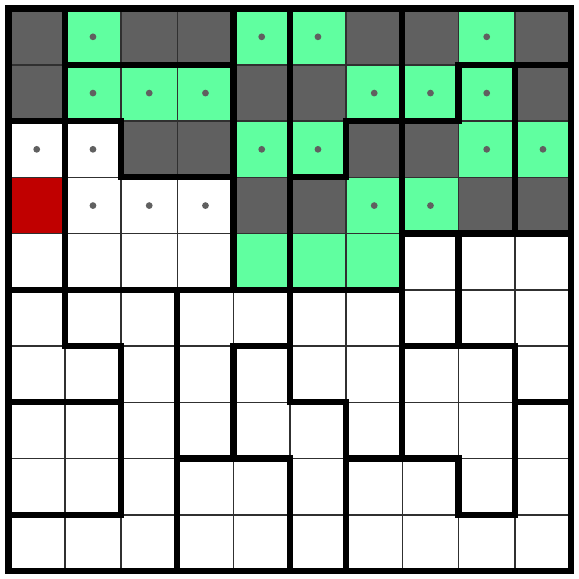}
    \caption{Partial solution with errors}
\end{subfigure}
\hfill
\begin{subfigure}[b]{0.3\textwidth}
    \includegraphics[width=\textwidth]{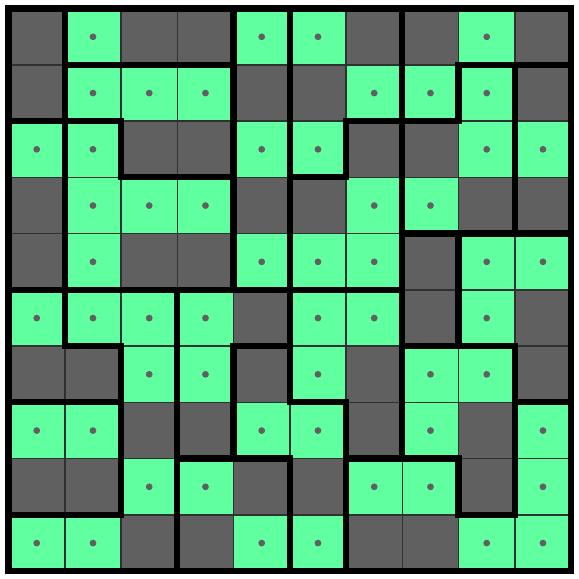}
    \caption{Completed}
\end{subfigure}
\caption{A norinori puzzle showing the solving process. Red highlights in (b) indicate cells where variety-specific constraints are violated; the verifier reports which rule is broken (e.g., ``Shaded cell has no adjacent shaded cell''), enabling targeted feedback in agentic mode.}
\label{fig:puzzle_example}
\end{figure}

\paragraph{Visualization}
Figure~\ref{fig:puzzle_example} shows a norinori puzzle at three stages: initial state, partial solution with constraint violations highlighted (red cells indicate which specific rule is broken), and completion.

\subsection{Dataset}

\paragraph{Full Database}
The complete dataset contains \textbf{\valFullPuzzles{} puzzles} across \valFullVarieties{} varieties, sourced from puzz.link, a Japanese puzzle community that aggregates puzzles from Twitter, personal blogs, and puzzle competitions. The dataset includes puzzles from \textbf{\valFullCreators{} unique creators} spanning multiple years of community activity. Each puzzle has:
\begin{itemize}
    \item \textbf{Verified unique solutions}: Every puzzle was validated via cspuz-solver2 to confirm exactly one solution exists
    \item \textbf{Step-by-step solutions}: Complete move sequences decomposed into \textit{required moves} and \textit{hint moves} (redundant moves that don't change the solution)
    \item \textbf{Source provenance}: Original puzzle URLs, author handles, and publication timestamps
\end{itemize}

\paragraph{Contamination}
Pencil puzzles are frequently shared online as puzz.link URLs, but \textit{solutions} are rarely published. The community shares puzzles for others to solve; posting solutions is considered poor etiquette. This means that while puzzle \textit{descriptions} may appear in training data, solution contamination is unlikely even for models trained on web data. This is a stronger contamination argument than mere source obscurity: the cultural norm of the puzzle community actively suppresses solution publication.

\paragraph{Golden Benchmark Set}
For evaluation, we curated a \textbf{\valGoldenPuzzles{}-puzzle golden set}:
\begin{itemize}
    \item \textbf{\valGoldenVarieties{} varieties}, selected as the most common varieties with verified solutions
    \item \textbf{\valPuzzlesPerType{} puzzles per type}, stratified by difficulty:
    \begin{itemize}
        \item \valPuzzlesPerTier{} short (fewest required moves within type)
        \item \valPuzzlesPerTier{} medium (middle quintile)
        \item \valPuzzlesPerTier{} long (most required moves within type)
    \end{itemize}
\end{itemize}

\paragraph{Dataset Summary}
Table~\ref{tab:dataset_summary} summarizes the key statistics across our dataset tiers.

\begin{table}[ht!]
\centering
\caption{Dataset summary statistics}
\label{tab:dataset_summary}
\begin{tabular}{lrrrr}
\toprule
 & Full Dataset & Golden 300 & Golden 30 & Golden 60 \\
\midrule
Puzzles & \valFullPuzzles{} & \valGoldenPuzzles{} & \valGoldenThirtyPuzzles{} & \valGoldenSixtyPuzzles{} \\
Varieties & \valFullVarieties{} & \valGoldenVarieties{} & \valGoldenThirtyVarieties{} & \valGoldenVarieties{} \\
Unique creators & \valFullCreators{} & \valGoldenCreators{} & \valGoldenThirtyCreators{} & -- \\
Median moves & \valFullMedianMoves{} & \valGoldenMedianMoves{} & \valGoldenThirtyMedianMoves{} & -- \\
Median area & \valFullMedianArea{} & \valGoldenMedianArea{} & \valGoldenMedianArea{} & -- \\
SAT-verified & 100\% & 100\% & 100\% & 100\% \\
\bottomrule
\end{tabular}
\end{table}

\subsection{Python Library}

We implemented \texttt{pencil-puzzle-bench} as a Python package:

\paragraph{Core API}
The \texttt{Puzzle} class wraps the pzprjs engine with a simple Python interface:
\begin{lstlisting}
from ppbench import Puzzle, load_dataset

puzzle = Puzzle.from_url("http://puzz.link/p?sudoku/...")
puzzle.send_move("mouse,left,3,5")  # Make a move
violations = puzzle.check()          # [] if valid
if puzzle.is_complete():
    print("Solved!")
\end{lstlisting}

\paragraph{Dataset Loading}
The golden benchmark set ships with the package:
\begin{lstlisting}
puzzles = load_dataset("golden")     # 300 puzzles
puzzles = load_dataset("golden_30")  # 30-puzzle subset
\end{lstlisting}

The full \valFullPuzzles{} puzzle dataset includes step-level solutions suitable for Gym-compatible RL training~\cite{brockman2016gym}.

\subsection{Puzzle Varieties}

Table~\ref{tab:puzzle_types} shows the \numpuzzletypes{} varieties in the evaluation set, ordered by solve rate.

\begin{table}[ht!]
\centering
\caption{Puzzle varieties by difficulty (best single-model solve rate across all strategies)}
\label{tab:puzzle_types}
\begin{tabular}{lrr}
\toprule
Variety & Best Model & Solve Rate \\
\midrule
norinori & gpt-5.2@xhigh & 100.0\% \\
shikaku & gpt-5.2@xhigh & 86.7\% \\
lits & gpt-5.2@xhigh & 73.3\% \\
hitori & claude-opus-4-6@thinking & 66.7\% \\
lightup & claude-opus-4-6@thinking & 66.7\% \\
mashu & gpt-5.2@xhigh & 66.7\% \\
tapa & gpt-5.2@xhigh & 66.7\% \\
firefly & gpt-5.2@xhigh & 46.7\% \\
nurimisaki & claude-opus-4-6@thinking & 40.0\% \\
yajilin & gpt-5.2@xhigh & 40.0\% \\
nurikabe & claude-opus-4-6@thinking & 33.3\% \\
slither & gpt-5.2@high & 33.3\% \\
sudoku & gpt-5.2@xhigh & 33.3\% \\
kurodoko & gpt-5.2@xhigh & 26.7\% \\
nurimaze & claude-opus-4-6@thinking & 26.7\% \\
sashigane$^*$ & gpt-5.2@xhigh & 20.0\% \\
country & gemini-3.1-pro & 13.3\% \\
dbchoco & gpt-5.2@xhigh & 6.7\% \\
heyawake$^*$ & gpt-5.2@xhigh & 6.7\% \\
shakashaka$^*$ & gpt-5.2@xhigh & 6.7\% \\
\bottomrule

\end{tabular}

\smallskip
\noindent\small{$^*$Solved only through agentic iteration (0\% in direct-ask evaluation).}
\end{table}

\section{Evaluation Methodology}
\label{sec:methodology}

\subsection{Strategies}

\paragraph{Direct Ask (Single-Shot)}
Model receives puzzle state and must output complete solution as JSON move list. One inference call.

\paragraph{Agentic (Multi-Turn with Iterative Verification)}
Model has access to tools: \texttt{make\_move}, \texttt{check\_board}, \texttt{reset\_puzzle}. The model iteratively makes moves, checks the board for constraint violations, and course-corrects based on the specific error feedback. This process continues until the puzzle is solved or the model gives up.

\medskip
\noindent\textbf{Move Format:} Moves are coordinate-based strings like \texttt{mouse,left,3,5} (left-click at column 3, row 5) or \texttt{mouse,right,1,2} (right-click for secondary state). Line-drawing puzzles use drag syntax: \texttt{mouse,left,1,1,1,5}. This interface mirrors the puzz.link web UI.

\medskip
\noindent\textit{Note:} Both strategies are \textbf{baselines}, not optimized attempts. We did not perform prompt engineering or hill-climbing. The goal is to measure where models are, not to maximize scores.

\subsection{Agentic Evaluation Design}

Of \nummodels{} models, \valAgenticExcludedModels{} were excluded from agentic evaluation because they lack the tool-calling capability required by the harness. The remaining \valAgenticTotalModels{} were evaluated agentically: \valAgenticBaselineModels{} on a \textbf{\valGoldenThirtyPuzzles{}-puzzle baseline} (\valGoldenThirtyVarieties{} types: yajilin, sashigane, lits, lightup), and \valAgenticExpandedModels{} top models (GPT-5.2@xhigh, Claude Opus 4.6@thinking, and Gemini 3.1 Pro) on an \textbf{expanded \valGoldenSixtyPuzzles{}-puzzle set} across all \valGoldenVarieties{} varieties (\valExpandedPuzzlesPerType{} puzzles per variety). This expansion is what solved the \valNewlySolvedPhrase{}.

\paragraph{Agentic Scale}
Agentic evaluation is a long-context, many-turn process. Across \valAgenticRuns{} agentic runs (\valAgenticBaselineModels{} models on the \valGoldenThirtyPuzzles{}-puzzle baseline, plus \valAgenticExpandedModels{} top models on the expanded \valGoldenSixtyPuzzles{}-puzzle set):
\begin{itemize}
    \item \textbf{Turns per attempt}: mean \valAgenticMeanTurns{}, median \valAgenticMedianTurns{}, P90 = \valAgenticPNinetyTurns{}, max \valAgenticMaxTurns{}
    \item \textbf{Duration per attempt}: mean \valAgenticMeanDuration{} min, median \valAgenticMedianDuration{} min, P90 = \valAgenticPNinetyDuration{} min, max \valAgenticMaxDuration{} hours
\end{itemize}

These are not brief tool calls; models sustain solving attempts over extended multi-turn conversations, maintaining context across dozens to hundreds of reasoning-action-feedback loops. This makes \benchmark{} a demanding test of long-context utilization and sustained reasoning, not just single-turn problem solving.

\subsection{Models}

We evaluate \nummodels{} models across \valNumProviders{} providers:
\begin{itemize}
    \item \textbf{OpenAI} (\valOpenAIConfigs{}): GPT-5.2 (none/low/medium/high/xhigh), GPT-5.2-pro, GPT-5.1@medium, GPT-5@medium, GPT-4.1, GPT-4o, o1, o3, GPT-3.5-turbo, GPT-OSS-120B
    \item \textbf{Anthropic} (\valAnthropicConfigs{}): Claude Sonnet 4.5, Claude Sonnet 4.5@thinking, Claude Sonnet 4.6, Claude Sonnet 4.6@thinking, Claude Sonnet 4.6-1m, Claude Opus 4.5 (high), Claude Opus 4.5@thinking, Claude Opus 4.6 (high), Claude Opus 4.6@thinking, Claude Opus 4.6@max, Claude Opus 4.6-1m
    \item \textbf{Google} (\valGoogleConfigs{}): Gemini 3.1 Pro, Gemini 3 Pro (default/minimal/high), Gemini 3 Flash (minimal/high), Gemini 2.5 Pro
    \item \textbf{xAI} (\valXAIConfigs{}): Grok 4.1 Fast, Grok 4.1 Fast Reasoning, Grok Code Fast
    \item \textbf{DeepSeek} (\valDeepSeekConfigs{}): V3.2, V3.2-speciale
    \item \textbf{Qwen} (\valQwenConfigs{}): Qwen3.5-397B, Qwen3-235B, Qwen3-Coder, Qwen3-Next-80B, Qwen3-VL-235B
    \item \textbf{Mistral} (\valMistralConfigs{}): Mistral Large, Devstral
    \item \textbf{Moonshot} (\valMoonshotConfigs{}): Kimi K2@thinking, Kimi K2.5
    \item \textbf{Minimax} (\valMinimaxConfigs{}): M2.1, M2.5
    \item \textbf{Zhipu} (\valZhipuConfigs{}): GLM-4.7, GLM-5
    \item \textbf{Xiaomi} (\valXiaomiConfigs{}): MiMo-v2
\end{itemize}

\section{Results}
\label{sec:results}

\subsection{Overall Performance}

Figure~\ref{fig:leaderboard} and Table~\ref{tab:main_results} show the top models ranked by their best success rate across either strategy. Complete results for all \nummodels{} models are in Table~\ref{tab:full_results} (Appendix C).

\begin{table}[ht!]
\centering
\small
\caption{Top \valTopNModels{} models by best success rate. Direct ask on \valGoldenPuzzles{} puzzles; agentic on \valGoldenThirtyPuzzles{}-puzzle baseline for most models, \valGoldenSixtyPuzzles{}-puzzle expanded set for top \valAgenticExpandedModels{}.}
\label{tab:main_results}
\begin{tabular}{lrr}
\toprule
Model & Direct Ask & Agentic \\
\midrule
gpt-5.2@xhigh & 27.0\% & \textbf{56.0\%} \\
claude-opus-4-6-1m & 0.0\% & \textbf{36.7\%} \\
gpt-5.2@high & 20.7\% & \textbf{36.7\%} \\
claude-opus-4-6@thinking & 27.3\% & \textbf{33.3\%} \\
gemini-3.1-pro & 20.0\% & \textbf{33.3\%} \\
claude-opus-4-6 & 0.3\% & \textbf{30.0\%} \\
claude-sonnet-4-6@thinking & 10.3\% & \textbf{26.7\%} \\
gpt-5.2-pro & 9.7\% & \textbf{26.7\%} \\
claude-opus-4-6@max & 0.3\% & \textbf{23.3\%} \\
claude-sonnet-4-6-1m & 0.3\% & \textbf{23.3\%} \\
gpt-5.2@medium & 9.3\% & \textbf{23.3\%} \\
claude-sonnet-4-6 & 0.3\% & \textbf{16.7\%} \\
gemini-3-pro@high & 3.3\% & \textbf{16.7\%} \\
gemini-3-pro & 4.3\% & \textbf{13.3\%} \\
gemini-3-pro@minimal & 4.0\% & \textbf{10.0\%} \\
\bottomrule

\end{tabular}
\end{table}

\subsection{The Agentic Gap}
\label{sec:agentic}

The agentic strategy, where models iteratively make moves, check the board for constraint violations, and course-correct, reveals a striking pattern: \textbf{models with weak direct-ask performance benefit disproportionately from agentic iteration}. Table~\ref{tab:agentic_uplift} shows the top 10 models by agentic success rate with uplift, and Table~\ref{tab:reasoning_scaling} shows GPT-5.2's performance across reasoning effort levels.

\begin{table}[ht!]
\centering
\small
\caption{Top 10 models by agentic success rate, with uplift ($\Delta$pp = agentic $-$ direct) and cost per attempt.}
\label{tab:agentic_uplift}
\begin{tabular}{lrrrr}
\toprule
Model & Direct & Agentic & $\Delta$pp & Cost/Attempt \\
\midrule
gpt-5.2@xhigh & 27.0\% & 56.0\% & +29.0 & \$9.74 \\
claude-opus-4-6-1m & 0.0\% & 36.7\% & +36.7 & \$14.11 \\
gpt-5.2@high & 20.7\% & 36.7\% & +16.0 & \$7.30 \\
claude-opus-4-6@thinking & 27.3\% & 33.3\% & +6.0 & \$6.24 \\
gemini-3.1-pro & 20.0\% & 33.3\% & +13.3 & \$14.36 \\
claude-opus-4-6 & 0.3\% & 30.0\% & +29.7 & \$10.89 \\
claude-sonnet-4-6@thinking & 10.3\% & 26.7\% & +16.3 & \$3.94 \\
gpt-5.2-pro & 9.7\% & 26.7\% & +17.0 & \$41.52 \\
claude-opus-4-6@max & 0.3\% & 23.3\% & +23.0 & \$10.87 \\
claude-sonnet-4-6-1m & 0.3\% & 23.3\% & +23.0 & \$169.27 \\
\bottomrule

\end{tabular}
\end{table}

\begin{table}[ht!]
\centering
\small
\caption{GPT-5.2 performance by reasoning effort level. Direct ask on \valGoldenPuzzles{} puzzles; agentic on \valGoldenThirtyPuzzles{}-puzzle baseline.}
\label{tab:reasoning_scaling}
\begin{tabular}{lrrr}
\toprule
Model & Direct & Agentic & Uplift \\
\midrule
gpt-5.2 & 0.33\% & 0.00\% & 1$\times$ \\
gpt-5.2@low & 2.3\% & 10.0\% & 7$\times$ \\
gpt-5.2@medium & 9.3\% & 23.3\% & 28$\times$ \\
gpt-5.2@high & 20.7\% & 36.7\% & 62$\times$ \\
gpt-5.2@xhigh & 27.0\% & 56.0\% & 81$\times$ \\
\bottomrule

\end{tabular}
\end{table}

Claude Opus 4.6 at default effort (no extended thinking) achieves \valClaudeOpusAgentic{}\% agentic success despite only \valClaudeOpusDirect{}\% direct-ask performance (\valClaudeOpusMatchedDelta{} percentage points on the \valClaudeOpusMatchedN{} puzzles evaluated in both modes). This is the most extreme case: the model lacks internal chain-of-thought reasoning, so agentic iteration substantially compensates. Even with extended thinking enabled, the gap persists: Claude Opus 4.6@thinking scores \valClaudeOpusThinkingDirect{}\% direct-ask but still gains \valClaudeOpusThinkingMatchedDelta{}pp from agentic iteration (\valClaudeOpusThinkingMatchedDirect{}\% $\to$ \valClaudeOpusThinkingMatchedAgentic{}\% on the same \valClaudeOpusThinkingMatchedN{} puzzles). GPT-5.2@xhigh, which already achieves \valGptXhighDirect{}\% direct-ask, gains \valGptXhighMatchedDelta{}pp from agentic iteration (\valGptXhighMatchedDirect{}\% $\to$ \valGptXhighMatchedAgentic{}\% on the same \valGptXhighMatchedN{} puzzles). This suggests that reasoning depth and agentic iteration are \textbf{two distinct axes of capability}: the agentic gap ranges from +\valClaudeOpusThinkingMatchedDelta{}pp for models with strong reasoning to +\valClaudeOpusMatchedDelta{}pp for models without, and both benefit from iterative verification and course-correction.

\paragraph{Extended Agentic Evaluation}
The initial \valGoldenThirtyPuzzles{}-puzzle agentic evaluation covered \valGoldenThirtyVarieties{} varieties. We then ran an expanded \valGoldenSixtyPuzzles{}-puzzle evaluation across all \valGoldenVarieties{} varieties for the top \valAgenticExpandedModels{} models (GPT-5.2@xhigh, Claude Opus 4.6@thinking, and Gemini 3.1 Pro). Infrastructure error rates (timeouts, API failures) differed between models: GPT-5.2@xhigh had a \valGptXhighErrThirty{}\% error rate on the baseline set vs.\ \valGptXhighErrSixty{}\% on the expanded puzzles, while Claude Opus 4.6@thinking showed \valClaudeOpusErrThirty{}\% and \valClaudeOpusErrSixty{}\% respectively, suggesting different failure modes across models.

\paragraph{Claude's Arc}
At the same configuration (default effort, no extended thinking), agentic performance improves across Claude generations: Sonnet 4.5 (\valClaudeSonnetDirect{}\% / \valClaudeSonnetAgentic{}\%), Opus 4.5 (\valClaudeOpusFourFiveDirect{}\% / \valClaudeOpusFourFiveAgentic{}\%), Opus 4.6 (\valClaudeOpusDirect{}\% / \valClaudeOpusAgentic{}\%). Without extended thinking, the jump from Opus 4.5 to 4.6 is entirely in agentic mode. With extended thinking enabled, both axes improve: Opus 4.5@thinking (\valClaudeOpusFourFiveThinkingDirect{}\% / \valClaudeOpusFourFiveThinkingAgentic{}\%) to Opus 4.6@thinking (\valClaudeOpusThinkingDirect{}\% / \valClaudeOpusThinkingAgentic{}\%).

\paragraph{Higher Effort Does Not Always Help}
Among non-thinking configurations, Claude Opus 4.6 at default effort (high) outperforms Opus 4.6@max (\valClaudeOpusAgentic{}\% vs \valClaudeOpusMaxAgentic{}\% on the baseline \valGoldenThirtyPuzzles{}-puzzle set). Enabling extended thinking (Opus 4.6@thinking, \valClaudeOpusThinkingAgentic{}\% on the expanded set) does not surpass the default effort configuration on the baseline puzzles, suggesting that the relationship between reasoning depth and agentic effectiveness is non-monotonic.

\paragraph{Agentic Iteration Solves Previously Unsolved Varieties}
The expanded agentic evaluation solved \valNewlySolvedPhrase{}; all \numpuzzletypes{} varieties have now been solved at least once. This demonstrates that iterative verification and course-correction can unlock capabilities not achieved by any model in single-shot mode.

\subsection{Reasoning Effort Scaling}

\begin{figure}[ht!]
\centering
\includegraphics[width=0.7\textwidth]{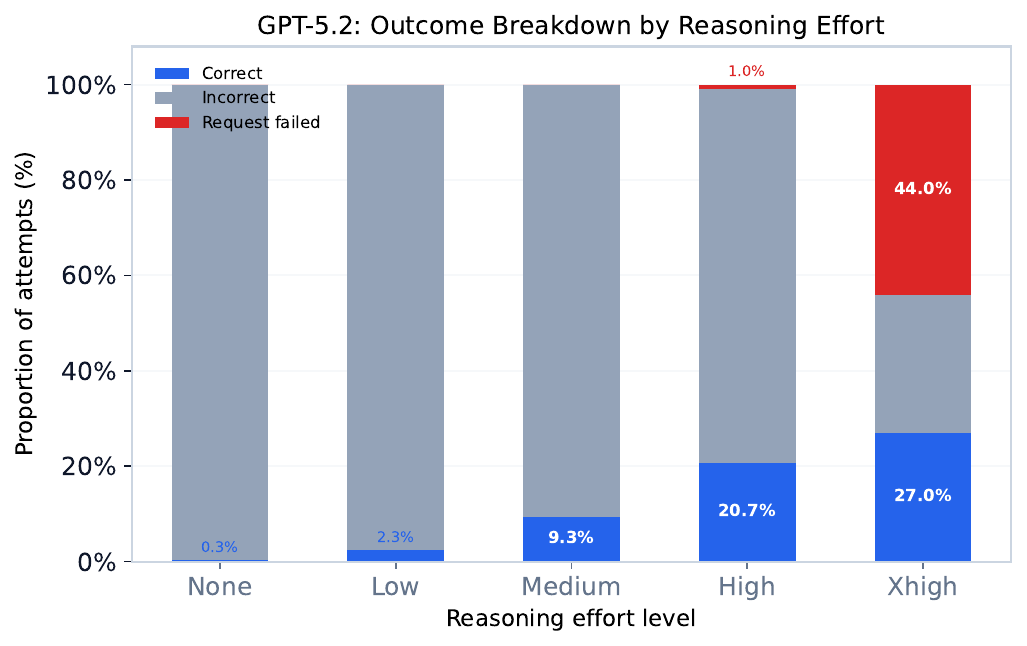}
\caption{GPT-5.2 outcome breakdown by reasoning effort level (direct ask, \valGoldenPuzzles{} puzzles). Each bar sums to 100\%. Correct answers (blue) improve \valReasoningScaling{}$\times$ from none to xhigh, but at xhigh, \valXhighErrorRate{}\% of requests fail before returning a response (red), revealing a sharp reliability/capability tradeoff.}
\label{fig:reasoning_scaling}
\end{figure}

Figure~\ref{fig:reasoning_scaling} shows GPT-5.2's outcome breakdown across reasoning effort levels. Each attempt has one of three outcomes: \textit{correct} (puzzle solved), \textit{incorrect} (model returned a response but the solution was wrong), or \textit{request failed} (infrastructure timeout or API error before any response was returned).

\subsection{Model Capability Over Time}

\begin{figure}[ht!]
\centering
\includegraphics[width=0.95\textwidth]{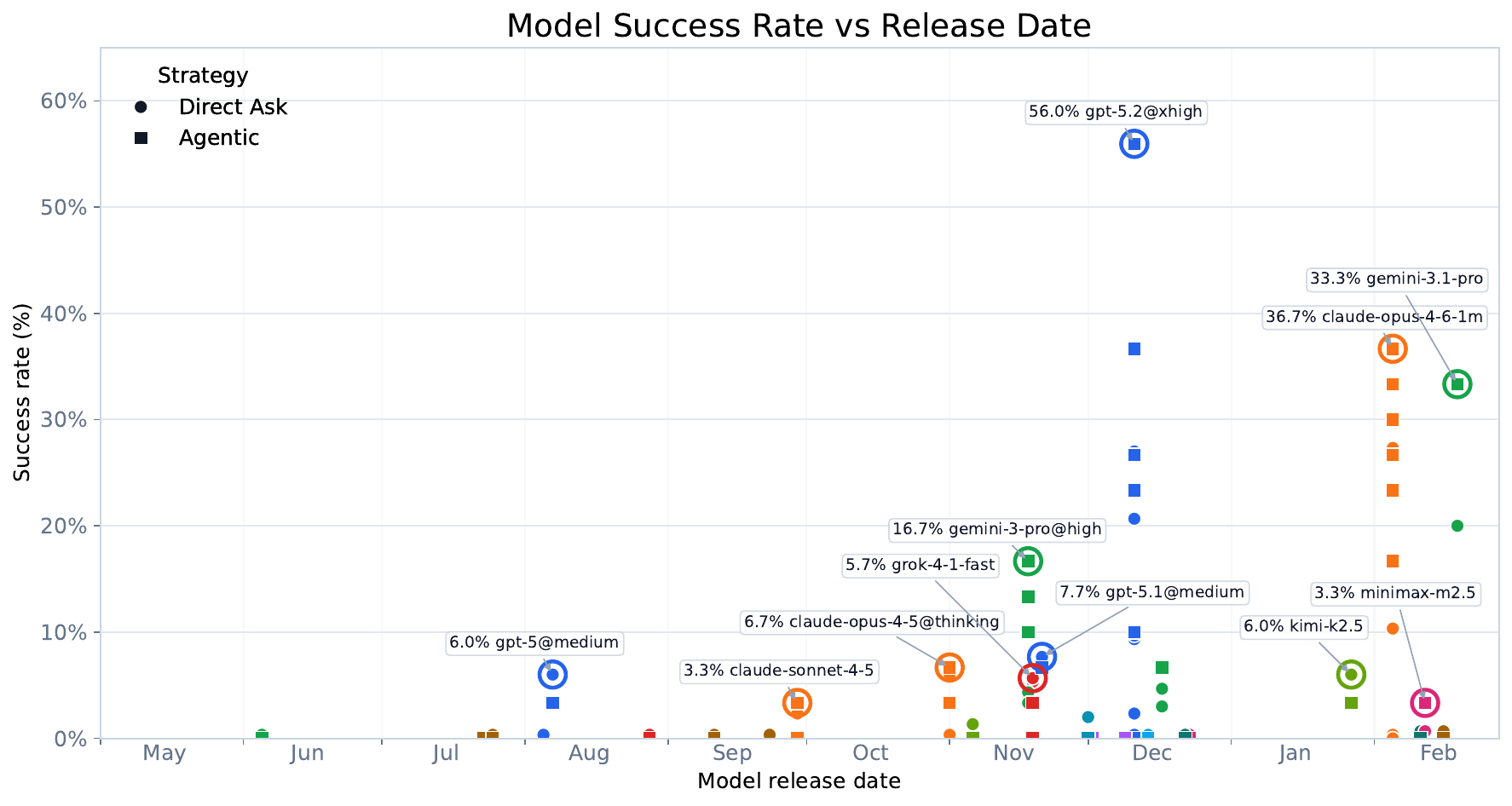}
\caption{Model success rates over time with frontier model release dates annotated. The progression shows both generational improvement within model families and the gap between frontier and non-frontier models.}
\label{fig:model_timeline}
\end{figure}

Figure~\ref{fig:model_timeline} shows model release dates plotted against success rates, illustrating the pace of capability improvement. At medium reasoning effort, we observe steady improvement across GPT-5 generations: GPT-5.0 (\valGptFiveZeroDirect{}\%), GPT-5.1 (\valGptFiveOneDirect{}\%), GPT-5.2 (\valGptFiveTwoDirect{}\%). A breakaway in capability is visible since late 2025, with dramatic improvements in the most recent three months. However, even the best models reach only \valGptXhighDirect{}\% success in direct-ask mode, and many varieties and larger puzzles remain unsolved across all models. The extended timeline in Figure~\ref{fig:full_timeline} (Appendix E) shows that models released before late 2024 achieve 0\% or near-0\% solve rates, confirming that this capability is entirely new.

\subsection{Cost Analysis}

\begin{figure}[ht!]
\centering
\includegraphics[width=0.8\textwidth]{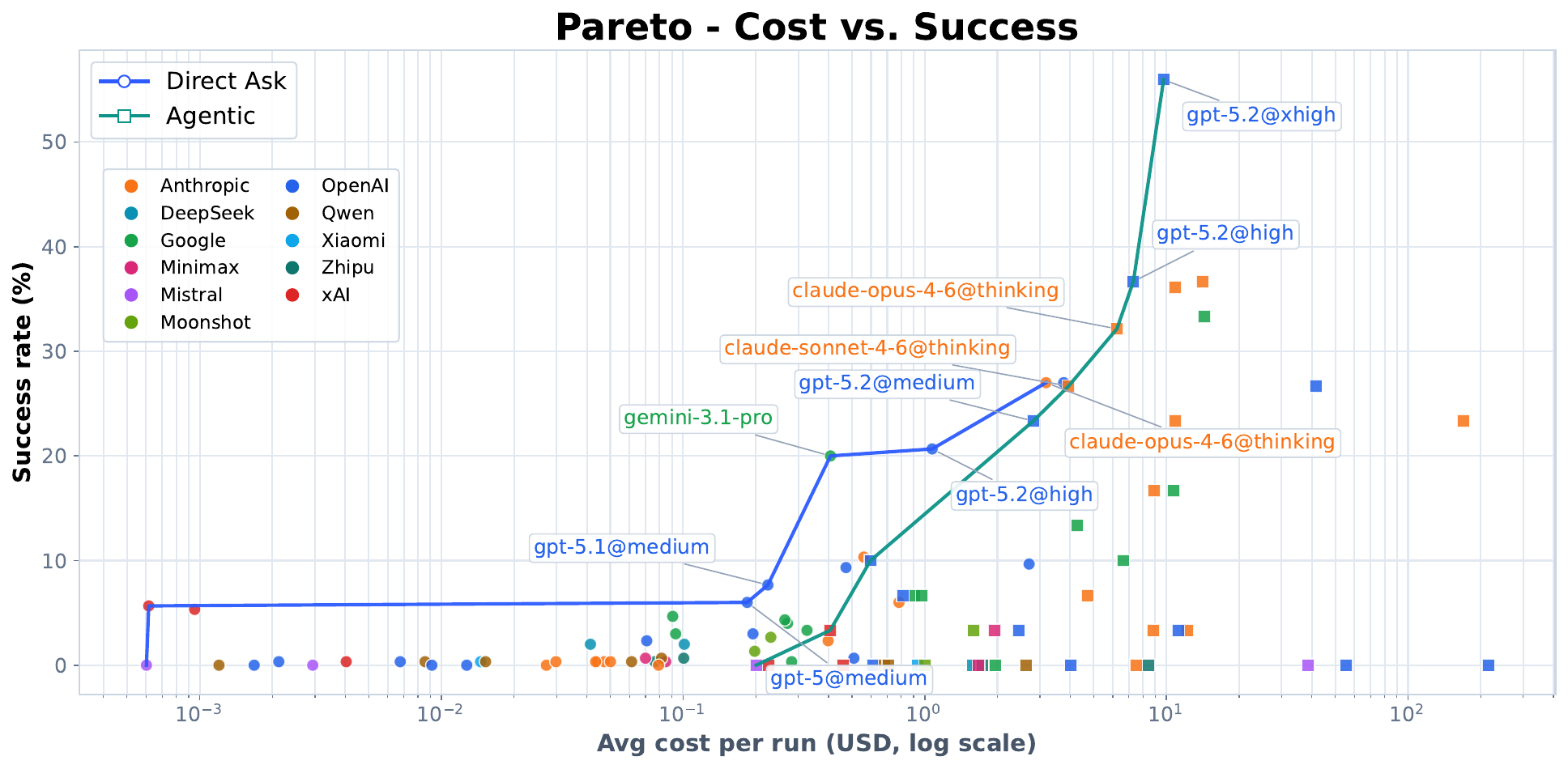}
\caption{Cost per success vs.\ success rate across models. The Pareto frontier (lower-right) shows the cost-efficiency tradeoff: Grok 4.1 Fast achieves the lowest cost per success (\$\valLowestCostPerSuccess{}) while GPT-5.2@xhigh achieves the highest success rate (\valGptXhighAgentic{}\%).}
\label{fig:pareto}
\end{figure}

Total recorded benchmark cost: approximately \$\valTotalCost{} across \valTotalRuns{} runs, computed from per-request token usage in the published artifacts.\footnote{These are artifact-recorded costs. Actual provider costs are higher than recorded because when agentic runs terminate with errors, token usage from completed turns before the error is not recorded.} All run records (model, puzzle, strategy, outcome, cost, duration) are published in the \texttt{runs.jsonl} artifact on HuggingFace.\footnote{\url{https://huggingface.co/datasets/approximatelabs/pencil-puzzle-bench}} Figure~\ref{fig:pareto} visualizes the cost-success Pareto frontier.

Cost-per-success varies by \textbf{\valCostVariation{}$\times$} across models.

\section{Analysis}
\label{sec:analysis}

\subsection{What Makes a Puzzle Hard?}
\label{sec:difficulty}

Our benchmark stratifies puzzles into short, medium, and long tiers by \textit{number of required moves} within each type. But is move count a good proxy for difficulty? After analysis, we propose \textbf{solution compression ratio} as a better difficulty measure.

\paragraph{Move Count is a Weak Predictor}
Regressing solve rate on move count yields $R^2_{\text{adj}} = \valMoveCountRsq{}$; move count explains only \valMoveCountRsq{} of variance in solve rate. Puzzle variety alone ($\eta^2 = \valTypeEtaSq{}$) explains \valTypeVsMovesFactor{}$\times$ more variance than move count.

\paragraph{Solution Compressibility Predicts Difficulty}
The \textbf{compression ratio} of the solution move sequence (zlib-compressed size / raw size) achieves $R^2_{\text{adj}} = \valCompressionRsq{}$, \valCompressionVsMovesFactor{}$\times$ more predictive than move count. The intuition: solutions with high compression ratios contain repetitive, structured patterns that models can generalize from partial progress. Low compression ratios indicate high-entropy solutions requiring genuinely novel decisions at each step.

\paragraph{Why Area Doesn't Add Much}
Adding grid area improves the model only modestly (compression + area + log(area): $R^2_{\text{adj}} = \valCompressionAreaRsq{}$ vs.\ compression alone: \valCompressionRsq{}). This is likely because area is confounded with variety; different varieties tend to use characteristic grid sizes. Within a variety, puzzles often cluster in a narrow size range (the Golden 300 has a median area of \valGoldenMedianArea{} for nearly every variety), so area provides little additional signal once variety is controlled for. Adding variety dummies to the best three features yields $R^2_{\text{adj}} = \valFullModelRsq{}$, confirming that variety captures additional difficulty structure beyond compressibility.

\subsection{Infrastructure Scaling Limits}

The \valXhighErrorRate{}\% request failure rate at xhigh reasoning effort is not random network failure. Error duration analysis reveals \valErrorClusterPct{}\% of failures occur between \valErrorClusterLoHr{}--\valErrorClusterHiHr{} hours, with \valPeakErrorPct{}\% clustering in the \valPeakLoHr{}--\valPeakHiHr{} hour range, suggesting a systematic server-side timeout around \valTimeoutEstLo{}--\valTimeoutEstHi{} hours. These are not transient errors but systematic limits on how long a single inference request can run.

The tradeoff is stark (Figure~\ref{fig:reasoning_scaling}): xhigh gains +\valXhighGainOverHigh{}pp over high in direct ask (\valGptHighDirect{}\% $\rightarrow$ \valGptXhighDirect{}\%), with corresponding gains in agentic mode (\valGptHighAgentic{}\% $\rightarrow$ \valGptXhighAgentic{}\%). But at xhigh, \valXhighErrorRate{}\% of requests fail before returning any response. The marginal capability gain is real, but reliability collapses.

\section{Discussion}
\label{sec:discussion}

\paragraph{Two Axes of Capability}
Our results reveal two distinct axes along which models can improve at puzzle solving: \textit{reasoning depth} (controlled by effort levels and extended thinking) and \textit{agentic iteration} (controlled by iterative verification and course-correction). These axes are complementary, not redundant. The agentic gap is largest for models without internal reasoning: Claude Opus 4.6 (no extended thinking) has near-zero direct-ask performance but reaches \valClaudeOpusAgentic{}\% agentically. With extended thinking enabled, Opus 4.6@thinking achieves \valClaudeOpusThinkingDirect{}\% direct-ask---yet still gains \valClaudeOpusThinkingMatchedDelta{}pp from agentic iteration (\valClaudeOpusThinkingMatchedDirect{}\% $\to$ \valClaudeOpusThinkingMatchedAgentic{}\% on the same \valClaudeOpusThinkingMatchedN{} puzzles). GPT-5.2@xhigh, the strongest single-shot reasoner at \valGptXhighDirect{}\%, gains a further +\valGptXhighMatchedDelta{}pp agentically (\valGptXhighAgentic{}\%). The strongest results combine both axes: deep reasoning \textit{and} iterative verification. The expanded agentic evaluation solved \valNewlySolvedPhrase{} that no model solved in direct-ask mode, demonstrating that agentic iteration provides capabilities beyond what single-shot reasoning alone achieves.

\paragraph{A Grounding Example}
To contextualize these results: the best model at maximum reasoning effort (GPT-5.2@xhigh) solves only \valSudokuBestRate{}\% of Sudoku puzzles, a variety familiar to most readers. This underscores that even well-known constraint-satisfaction problems remain genuinely challenging for frontier models, despite Sudoku being one of the most commonly encountered varieties in training data.

\paragraph{Limitations}
Models received ASCII text-based board representations. In agentic mode, models also had access to a \texttt{render\_board\_as\_svg()} tool that returns an SVG text representation of the current board state, which some models called for additional spatial context. Rasterized pixel images were not used. Varieties requiring strong visual-spatial reasoning may benefit from multimodal approaches using rendered images. The agentic evaluation used a \textbf{\valGoldenThirtyPuzzles{}-puzzle subset} (\valGoldenThirtyVarieties{} varieties: yajilin, sashigane, lits, lightup, selected to span a range of difficulty levels and elicit meaningful agentic iteration) for most models, with an expanded \textbf{\valGoldenSixtyPuzzles{}-puzzle evaluation} (\valExpandedPuzzlesPerType{} puzzles per variety across all \valGoldenVarieties{} varieties, selected by difficulty tier) for the top \valAgenticExpandedModels{} models. Both strategies are baselines without prompt optimization.

All success rates are point estimates from single evaluation runs without confidence intervals or repeated trials. With \valPuzzlesPerType{} puzzles per variety, a single additional solve changes per-variety rates by \valPerTypePPChange{} percentage points. The agentic uplift figures are computed from small samples and should be interpreted with this caveat. No human performance baseline is established. Future work should include repeated trials, bootstrap confidence intervals, and human calibration studies.

\paragraph{Future Work}
The puzzle dataset, benchmark code, and step-level solution traces enable:
\begin{itemize}
    \item RL fine-tuning with step-level rewards, following the paradigm of process supervision~\cite{lightman2023prm,zhang2025prm} and reasoning via RL~\cite{deepseek2025r1}
    \item Process reward model training using per-move, per-constraint verification signals
    \item Curriculum learning across difficulty levels using compression ratio as a principled difficulty metric
    \item Multimodal evaluation using the SVG renderer, where models could ``see'' the puzzle board as a rendered image rather than parsing ASCII text
\end{itemize}

\section{Conclusion}
\label{sec:conclusion}

We introduced \benchmark{}, a dataset of \valFullPuzzles{} logic puzzles across \valFullVarieties{} varieties with step-level solution traces, and an evaluation benchmark of \numpuzzles{} puzzles across \numpuzzletypes{} varieties with programmatic step-level verification, where every intermediate board state can be checked against variety-specific constraints. Our evaluation of \nummodels{} models in two modes (direct ask and agentic) reveals two distinct axes of improvement: reasoning effort scaling (\valReasoningScaling{}$\times$ from GPT-5.2 none to xhigh) and agentic iteration (up to +\valClaudeOpusMatchedDelta{}pp for models without extended thinking, and +\valClaudeOpusThinkingMatchedDelta{}pp even for strong reasoning models). The strongest results combine both axes: GPT-5.2@xhigh achieves \valGptXhighAgentic{}\% in agentic mode. Agentic evaluation solved \valNewlySolvedPhrase{}; all \numpuzzletypes{} varieties have now been solved at least once, though \valUnsolvedPostAgentic{}\% of individual puzzles remain unsolved by any model.

The step-level verification infrastructure provides a foundation for future work on process supervision, step-level reinforcement learning, and curriculum learning for reasoning.


\paragraph{Acknowledgments}
This paper was written with heavy use of AI coding agents (Claude Code, Codex) and conversational AI assistants (ChatGPT, Claude). These tools were used extensively throughout the research, implementation, data analysis, and writing process. Any errors or inaccuracies in this work are the sole responsibility of the author.


\bibliographystyle{ppbench}
\bibliography{references}

@misc{benchmarkinadequacies2024,
      title={Inadequacies of Large Language Model Benchmarks in the Era of Generative Artificial Intelligence}, 
      author={Timothy R. McIntosh and Teo Susnjak and Nalin Arachchilage and Tong Liu and Paul Watters and Malka N. Halgamuge},
      year={2024},
      eprint={2402.09880},
      archivePrefix={arXiv},
      primaryClass={cs.AI},
      doi={https://doi.org/10.1109/TAI.2025.3569516},
      url={https://arxiv.org/abs/2402.09880}, 
}

@misc{besta2023got,
      title={Graph of Thoughts: Solving Elaborate Problems with Large Language Models}, 
      author={Maciej Besta and Nils Blach and Ales Kubicek and Robert Gerstenberger and Michal Podstawski and Lukas Gianinazzi and Joanna Gajda and Tomasz Lehmann and Hubert Niewiadomski and Piotr Nyczyk and Torsten Hoefler},
      year={2024},
      eprint={2308.09687},
      archivePrefix={arXiv},
      primaryClass={cs.CL},
      doi={https://doi.org/10.1609/aaai.v38i16.29720},
      url={https://arxiv.org/abs/2308.09687}, 
}

@misc{brockman2016gym,
      title={OpenAI Gym}, 
      author={Greg Brockman and Vicki Cheung and Ludwig Pettersson and Jonas Schneider and John Schulman and Jie Tang and Wojciech Zaremba},
      year={2016},
      eprint={1606.01540},
      archivePrefix={arXiv},
      primaryClass={cs.LG},
      url={https://arxiv.org/abs/1606.01540}, 
}

@misc{chen2022pot,
      title={Program of Thoughts Prompting: Disentangling Computation from Reasoning for Numerical Reasoning Tasks}, 
      author={Wenhu Chen and Xueguang Ma and Xinyi Wang and William W. Cohen},
      year={2023},
      eprint={2211.12588},
      archivePrefix={arXiv},
      primaryClass={cs.CL},
      url={https://arxiv.org/abs/2211.12588}, 
}

@misc{chen2025enigmata,
      title={Enigmata: Scaling Logical Reasoning in Large Language Models with Synthetic Verifiable Puzzles}, 
      author={Jiangjie Chen and Qianyu He and Siyu Yuan and Aili Chen and Zhicheng Cai and Weinan Dai and Hongli Yu and Qiying Yu and Xuefeng Li and Jiaze Chen and Hao Zhou and Mingxuan Wang},
      year={2025},
      eprint={2505.19914},
      archivePrefix={arXiv},
      primaryClass={cs.CL},
      url={https://arxiv.org/abs/2505.19914}, 
}

@misc{chen2025finereason,
      title={FINEREASON: Evaluating and Improving LLMs' Deliberate Reasoning through Reflective Puzzle Solving}, 
      author={Guizhen Chen and Weiwen Xu and Hao Zhang and Hou Pong Chan and Chaoqun Liu and Lidong Bing and Deli Zhao and Anh Tuan Luu and Yu Rong},
      year={2025},
      eprint={2502.20238},
      archivePrefix={arXiv},
      primaryClass={cs.CL},
      url={https://arxiv.org/abs/2502.20238}, 
}

@misc{clark2018arc,
      title={Think you have Solved Question Answering? Try ARC, the AI2 Reasoning Challenge}, 
      author={Peter Clark and Isaac Cowhey and Oren Etzioni and Tushar Khot and Ashish Sabharwal and Carissa Schoenick and Oyvind Tafjord},
      year={2018},
      eprint={1803.05457},
      archivePrefix={arXiv},
      primaryClass={cs.AI},
      url={https://arxiv.org/abs/1803.05457}, 
}

@misc{clark2020ruletaker,
      title={Transformers as Soft Reasoners over Language}, 
      author={Peter Clark and Oyvind Tafjord and Kyle Richardson},
      year={2020},
      eprint={2002.05867},
      archivePrefix={arXiv},
      primaryClass={cs.CL},
      url={https://arxiv.org/abs/2002.05867}, 
}

@misc{cobbe2021gsm8k,
      title={Training Verifiers to Solve Math Word Problems}, 
      author={Karl Cobbe and Vineet Kosaraju and Mohammad Bavarian and Mark Chen and Heewoo Jun and Lukasz Kaiser and Matthias Plappert and Jerry Tworek and Jacob Hilton and Reiichiro Nakano and Christopher Hesse and John Schulman},
      year={2021},
      eprint={2110.14168},
      archivePrefix={arXiv},
      primaryClass={cs.LG},
      url={https://arxiv.org/abs/2110.14168}, 
}

@misc{deepseek2025r1,
      title={DeepSeek-R1: Incentivizing Reasoning Capability in LLMs via Reinforcement Learning}, 
      author={DeepSeek-AI and Daya Guo and Dejian Yang and Haowei Zhang and Junxiao Song and Peiyi Wang and Qihao Zhu and Runxin Xu and Ruoyu Zhang and Shirong Ma and Xiao Bi and Xiaokang Zhang and Xingkai Yu and Yu Wu and Z. F. Wu and Zhibin Gou and Zhihong Shao and Zhuoshu Li and Ziyi Gao and Aixin Liu and Bing Xue and Bingxuan Wang and Bochao Wu and Bei Feng and Chengda Lu and Chenggang Zhao and Chengqi Deng and Chenyu Zhang and Chong Ruan and Damai Dai and Deli Chen and Dongjie Ji and Erhang Li and Fangyun Lin and Fucong Dai and Fuli Luo and Guangbo Hao and Guanting Chen and Guowei Li and H. Zhang and Han Bao and Hanwei Xu and Haocheng Wang and Honghui Ding and Huajian Xin and Huazuo Gao and Hui Qu and Hui Li and Jianzhong Guo and Jiashi Li and Jiawei Wang and Jingchang Chen and Jingyang Yuan and Junjie Qiu and Junlong Li and J. L. Cai and Jiaqi Ni and Jian Liang and Jin Chen and Kai Dong and Kai Hu and Kaige Gao and Kang Guan and Kexin Huang and Kuai Yu and Lean Wang and Lecong Zhang and Liang Zhao and Litong Wang and Liyue Zhang and Lei Xu and Leyi Xia and Mingchuan Zhang and Minghua Zhang and Minghui Tang and Meng Li and Miaojun Wang and Mingming Li and Ning Tian and Panpan Huang and Peng Zhang and Qiancheng Wang and Qinyu Chen and Qiushi Du and Ruiqi Ge and Ruisong Zhang and Ruizhe Pan and Runji Wang and R. J. Chen and R. L. Jin and Ruyi Chen and Shanghao Lu and Shangyan Zhou and Shanhuang Chen and Shengfeng Ye and Shiyu Wang and Shuiping Yu and Shunfeng Zhou and Shuting Pan and S. S. Li and Shuang Zhou and Shaoqing Wu and Shengfeng Ye and Tao Yun and Tian Pei and Tianyu Sun and T. Wang and Wangding Zeng and Wanjia Zhao and Wen Liu and Wenfeng Liang and Wenjun Gao and Wenqin Yu and Wentao Zhang and W. L. Xiao and Wei An and Xiaodong Liu and Xiaohan Wang and Xiaokang Chen and Xiaotao Nie and Xin Cheng and Xin Liu and Xin Xie and Xingchao Liu and Xinyu Yang and Xinyuan Li and Xuecheng Su and Xuheng Lin and X. Q. Li and Xiangyue Jin and Xiaojin Shen and Xiaosha Chen and Xiaowen Sun and Xiaoxiang Wang and Xinnan Song and Xinyi Zhou and Xianzu Wang and Xinxia Shan and Y. K. Li and Y. Q. Wang and Y. X. Wei and Yang Zhang and Yanhong Xu and Yao Li and Yao Zhao and Yaofeng Sun and Yaohui Wang and Yi Yu and Yichao Zhang and Yifan Shi and Yiliang Xiong and Ying He and Yishi Piao and Yisong Wang and Yixuan Tan and Yiyang Ma and Yiyuan Liu and Yongqiang Guo and Yuan Ou and Yuduan Wang and Yue Gong and Yuheng Zou and Yujia He and Yunfan Xiong and Yuxiang Luo and Yuxiang You and Yuxuan Liu and Yuyang Zhou and Y. X. Zhu and Yanhong Xu and Yanping Huang and Yaohui Li and Yi Zheng and Yuchen Zhu and Yunxian Ma and Ying Tang and Yukun Zha and Yuting Yan and Z. Z. Ren and Zehui Ren and Zhangli Sha and Zhe Fu and Zhean Xu and Zhenda Xie and Zhengyan Zhang and Zhewen Hao and Zhicheng Ma and Zhigang Yan and Zhiyu Wu and Zihui Gu and Zijia Zhu and Zijun Liu and Zilin Li and Ziwei Xie and Ziyang Song and Zizheng Pan and Zhen Huang and Zhipeng Xu and Zhongyu Zhang and Zhen Zhang},
      year={2026},
      eprint={2501.12948},
      archivePrefix={arXiv},
      primaryClass={cs.CL},
      doi={https://doi.org/10.1038/s41586-025-09422-z},
      url={https://arxiv.org/abs/2501.12948}, 
}

@misc{deng2023contamination,
      title={Investigating Data Contamination in Modern Benchmarks for Large Language Models}, 
      author={Chunyuan Deng and Yilun Zhao and Xiangru Tang and Mark Gerstein and Arman Cohan},
      year={2024},
      eprint={2311.09783},
      archivePrefix={arXiv},
      primaryClass={cs.CL},
      url={https://arxiv.org/abs/2311.09783}, 
}

@misc{geva2021strategyqa,
      title={Did Aristotle Use a Laptop? A Question Answering Benchmark with Implicit Reasoning Strategies}, 
      author={Mor Geva and Daniel Khashabi and Elad Segal and Tushar Khot and Dan Roth and Jonathan Berant},
      year={2021},
      eprint={2101.02235},
      archivePrefix={arXiv},
      primaryClass={cs.CL},
      url={https://arxiv.org/abs/2101.02235}, 
}

@misc{giadikiaroglou2024puzzlesurvey,
      title={Puzzle Solving using Reasoning of Large Language Models: A Survey}, 
      author={Panagiotis Giadikiaroglou and Maria Lymperaiou and Giorgos Filandrianos and Giorgos Stamou},
      year={2024},
      eprint={2402.11291},
      archivePrefix={arXiv},
      primaryClass={cs.CL},
      doi={https://doi.org/10.18653/v1/2024.emnlp-main.646},
      url={https://arxiv.org/abs/2402.11291}, 
}

@misc{guo2024multiagents,
      title={Large Language Model based Multi-Agents: A Survey of Progress and Challenges}, 
      author={Taicheng Guo and Xiuying Chen and Yaqi Wang and Ruidi Chang and Shichao Pei and Nitesh V. Chawla and Olaf Wiest and Xiangliang Zhang},
      year={2024},
      eprint={2402.01680},
      archivePrefix={arXiv},
      primaryClass={cs.CL},
      url={https://arxiv.org/abs/2402.01680}, 
}

@misc{halder2025riddlebench,
      title={RiddleBench: A New Generative Reasoning Benchmark for LLMs}, 
      author={Deepon Halder and Alan Saji and Thanmay Jayakumar and Ratish Puduppully and Anoop Kunchukuttan and Raj Dabre},
      year={2025},
      eprint={2510.24932},
      archivePrefix={arXiv},
      primaryClass={cs.CL},
      url={https://arxiv.org/abs/2510.24932}, 
}

@misc{hendrycks2021math,
      title={Measuring Mathematical Problem Solving With the MATH Dataset}, 
      author={Dan Hendrycks and Collin Burns and Saurav Kadavath and Akul Arora and Steven Basart and Eric Tang and Dawn Song and Jacob Steinhardt},
      year={2021},
      eprint={2103.03874},
      archivePrefix={arXiv},
      primaryClass={cs.LG},
      url={https://arxiv.org/abs/2103.03874}, 
}

@misc{jimenez2024swebench,
      title={SWE-bench: Can Language Models Resolve Real-World GitHub Issues?}, 
      author={Carlos E. Jimenez and John Yang and Alexander Wettig and Shunyu Yao and Kexin Pei and Ofir Press and Karthik Narasimhan},
      year={2024},
      eprint={2310.06770},
      archivePrefix={arXiv},
      primaryClass={cs.CL},
      url={https://arxiv.org/abs/2310.06770}, 
}

@misc{ke2025llmreasoning,
      title={A Survey of Frontiers in LLM Reasoning: Inference Scaling, Learning to Reason, and Agentic Systems}, 
      author={Zixuan Ke and Fangkai Jiao and Yifei Ming and Xuan-Phi Nguyen and Austin Xu and Do Xuan Long and Minzhi Li and Chengwei Qin and Peifeng Wang and Silvio Savarese and Caiming Xiong and Shafiq Joty},
      year={2025},
      eprint={2504.09037},
      archivePrefix={arXiv},
      primaryClass={cs.AI},
      url={https://arxiv.org/abs/2504.09037}, 
}

@misc{li2024minesweeper,
      title={Assessing Logical Puzzle Solving in Large Language Models: Insights from a Minesweeper Case Study}, 
      author={Yinghao Li and Haorui Wang and Chao Zhang},
      year={2024},
      eprint={2311.07387},
      archivePrefix={arXiv},
      primaryClass={cs.CL},
      doi={https://doi.org/10.18653/v1/2024.naacl-long.4},
      url={https://arxiv.org/abs/2311.07387}, 
}

@misc{liang2025hardcorelogic,
      title={HardcoreLogic: Challenging Large Reasoning Models with Long-tail Logic Puzzle Games}, 
      author={Jingcong Liang and Shijun Wan and Xuehai Wu and Yitong Li and Qianglong Chen and Duyu Tang and Siyuan Wang and Zhongyu Wei},
      year={2025},
      eprint={2510.12563},
      archivePrefix={arXiv},
      primaryClass={cs.AI},
      url={https://arxiv.org/abs/2510.12563}, 
}

@misc{lightman2023prm,
      title={Let's Verify Step by Step}, 
      author={Hunter Lightman and Vineet Kosaraju and Yura Burda and Harri Edwards and Bowen Baker and Teddy Lee and Jan Leike and John Schulman and Ilya Sutskever and Karl Cobbe},
      year={2023},
      eprint={2305.20050},
      archivePrefix={arXiv},
      primaryClass={cs.LG},
      url={https://arxiv.org/abs/2305.20050}, 
}

@misc{lin2025zebralogic,
      title={ZebraLogic: On the Scaling Limits of LLMs for Logical Reasoning}, 
      author={Bill Yuchen Lin and Ronan Le Bras and Kyle Richardson and Ashish Sabharwal and Radha Poovendran and Peter Clark and Yejin Choi},
      year={2025},
      eprint={2502.01100},
      archivePrefix={arXiv},
      primaryClass={cs.AI},
      url={https://arxiv.org/abs/2502.01100}, 
}

@misc{madaan2023selfrefine,
      title={Self-Refine: Iterative Refinement with Self-Feedback}, 
      author={Aman Madaan and Niket Tandon and Prakhar Gupta and Skyler Hallinan and Luyu Gao and Sarah Wiegreffe and Uri Alon and Nouha Dziri and Shrimai Prabhumoye and Yiming Yang and Shashank Gupta and Bodhisattwa Prasad Majumder and Katherine Hermann and Sean Welleck and Amir Yazdanbakhsh and Peter Clark},
      year={2023},
      eprint={2303.17651},
      archivePrefix={arXiv},
      primaryClass={cs.CL},
      url={https://arxiv.org/abs/2303.17651}, 
}

@misc{nikolaev2025evolomino,
      title={Evolomino is NP-complete}, 
      author={Andrei V. Nikolaev},
      year={2025},
      eprint={2503.07611},
      archivePrefix={arXiv},
      primaryClass={cs.CC},
      doi={https://doi.org/10.33048/semi.2025.22.C05},
      url={https://arxiv.org/abs/2503.07611}, 
}

@misc{openai2024o1,
      title={OpenAI o1 System Card}, 
      author={OpenAI and : and Aaron Jaech and Adam Kalai and Adam Lerer and Adam Richardson and Ahmed El-Kishky and Aiden Low and Alec Helyar and Aleksander Madry and Alex Beutel and Alex Carney and Alex Iftimie and Alex Karpenko and Alex Tachard Passos and Alexander Neitz and Alexander Prokofiev and Alexander Wei and Allison Tam and Ally Bennett and Ananya Kumar and Andre Saraiva and Andrea Vallone and Andrew Duberstein and Andrew Kondrich and Andrey Mishchenko and Andy Applebaum and Angela Jiang and Ashvin Nair and Barret Zoph and Behrooz Ghorbani and Ben Rossen and Benjamin Sokolowsky and Boaz Barak and Bob McGrew and Borys Minaiev and Botao Hao and Bowen Baker and Brandon Houghton and Brandon McKinzie and Brydon Eastman and Camillo Lugaresi and Cary Bassin and Cary Hudson and Chak Ming Li and Charles de Bourcy and Chelsea Voss and Chen Shen and Chong Zhang and Chris Koch and Chris Orsinger and Christopher Hesse and Claudia Fischer and Clive Chan and Dan Roberts and Daniel Kappler and Daniel Levy and Daniel Selsam and David Dohan and David Farhi and David Mely and David Robinson and Dimitris Tsipras and Doug Li and Dragos Oprica and Eben Freeman and Eddie Zhang and Edmund Wong and Elizabeth Proehl and Enoch Cheung and Eric Mitchell and Eric Wallace and Erik Ritter and Evan Mays and Fan Wang and Felipe Petroski Such and Filippo Raso and Florencia Leoni and Foivos Tsimpourlas and Francis Song and Fred von Lohmann and Freddie Sulit and Geoff Salmon and Giambattista Parascandolo and Gildas Chabot and Grace Zhao and Greg Brockman and Guillaume Leclerc and Hadi Salman and Haiming Bao and Hao Sheng and Hart Andrin and Hessam Bagherinezhad and Hongyu Ren and Hunter Lightman and Hyung Won Chung and Ian Kivlichan and Ian O'Connell and Ian Osband and Ignasi Clavera Gilaberte and Ilge Akkaya and Ilya Kostrikov and Ilya Sutskever and Irina Kofman and Jakub Pachocki and James Lennon and Jason Wei and Jean Harb and Jerry Twore and Jiacheng Feng and Jiahui Yu and Jiayi Weng and Jie Tang and Jieqi Yu and Joaquin Quiñonero Candela and Joe Palermo and Joel Parish and Johannes Heidecke and John Hallman and John Rizzo and Jonathan Gordon and Jonathan Uesato and Jonathan Ward and Joost Huizinga and Julie Wang and Kai Chen and Kai Xiao and Karan Singhal and Karina Nguyen and Karl Cobbe and Katy Shi and Kayla Wood and Kendra Rimbach and Keren Gu-Lemberg and Kevin Liu and Kevin Lu and Kevin Stone and Kevin Yu and Lama Ahmad and Lauren Yang and Leo Liu and Leon Maksin and Leyton Ho and Liam Fedus and Lilian Weng and Linden Li and Lindsay McCallum and Lindsey Held and Lorenz Kuhn and Lukas Kondraciuk and Lukasz Kaiser and Luke Metz and Madelaine Boyd and Maja Trebacz and Manas Joglekar and Mark Chen and Marko Tintor and Mason Meyer and Matt Jones and Matt Kaufer and Max Schwarzer and Meghan Shah and Mehmet Yatbaz and Melody Y. Guan and Mengyuan Xu and Mengyuan Yan and Mia Glaese and Mianna Chen and Michael Lampe and Michael Malek and Michele Wang and Michelle Fradin and Mike McClay and Mikhail Pavlov and Miles Wang and Mingxuan Wang and Mira Murati and Mo Bavarian and Mostafa Rohaninejad and Nat McAleese and Neil Chowdhury and Neil Chowdhury and Nick Ryder and Nikolas Tezak and Noam Brown and Ofir Nachum and Oleg Boiko and Oleg Murk and Olivia Watkins and Patrick Chao and Paul Ashbourne and Pavel Izmailov and Peter Zhokhov and Rachel Dias and Rahul Arora and Randall Lin and Rapha Gontijo Lopes and Raz Gaon and Reah Miyara and Reimar Leike and Renny Hwang and Rhythm Garg and Robin Brown and Roshan James and Rui Shu and Ryan Cheu and Ryan Greene and Saachi Jain and Sam Altman and Sam Toizer and Sam Toyer and Samuel Miserendino and Sandhini Agarwal and Santiago Hernandez and Sasha Baker and Scott McKinney and Scottie Yan and Shengjia Zhao and Shengli Hu and Shibani Santurkar and Shraman Ray Chaudhuri and Shuyuan Zhang and Siyuan Fu and Spencer Papay and Steph Lin and Suchir Balaji and Suvansh Sanjeev and Szymon Sidor and Tal Broda and Aidan Clark and Tao Wang and Taylor Gordon and Ted Sanders and Tejal Patwardhan and Thibault Sottiaux and Thomas Degry and Thomas Dimson and Tianhao Zheng and Timur Garipov and Tom Stasi and Trapit Bansal and Trevor Creech and Troy Peterson and Tyna Eloundou and Valerie Qi and Vineet Kosaraju and Vinnie Monaco and Vitchyr Pong and Vlad Fomenko and Weiyi Zheng and Wenda Zhou and Wes McCabe and Wojciech Zaremba and Yann Dubois and Yinghai Lu and Yining Chen and Young Cha and Yu Bai and Yuchen He and Yuchen Zhang and Yunyun Wang and Zheng Shao and Zhuohan Li},
      year={2024},
      eprint={2412.16720},
      archivePrefix={arXiv},
      primaryClass={cs.AI},
      url={https://arxiv.org/abs/2412.16720}, 
}

@misc{park2023generativeagents,
      title={Generative Agents: Interactive Simulacra of Human Behavior}, 
      author={Joon Sung Park and Joseph C. O'Brien and Carrie J. Cai and Meredith Ringel Morris and Percy Liang and Michael S. Bernstein},
      year={2023},
      eprint={2304.03442},
      archivePrefix={arXiv},
      primaryClass={cs.HC},
      url={https://arxiv.org/abs/2304.03442}, 
}

@InProceedings{patil2025bfcl,
  title = 	 {The Berkeley Function Calling Leaderboard ({BFCL}): From Tool Use to Agentic Evaluation of Large Language Models},
  author =       {Patil, Shishir G and Mao, Huanzhi and Yan, Fanjia and Ji, Charlie Cheng-Jie and Suresh, Vishnu and Stoica, Ion and Gonzalez, Joseph E.},
  booktitle = 	 {Proceedings of the 42nd International Conference on Machine Learning},
  pages = 	 {48371--48392},
  year = 	 {2025},
  editor = 	 {Singh, Aarti and Fazel, Maryam and Hsu, Daniel and Lacoste-Julien, Simon and Berkenkamp, Felix and Maharaj, Tegan and Wagstaff, Kiri and Zhu, Jerry},
  volume = 	 {267},
  series = 	 {Proceedings of Machine Learning Research},
  month = 	 {13--19 Jul},
  publisher =    {PMLR},
  url = 	 {https://proceedings.mlr.press/v267/patil25a.html}
}

@misc{schick2023toolformer,
      title={Toolformer: Language Models Can Teach Themselves to Use Tools}, 
      author={Timo Schick and Jane Dwivedi-Yu and Roberto Dessì and Roberta Raileanu and Maria Lomeli and Luke Zettlemoyer and Nicola Cancedda and Thomas Scialom},
      year={2023},
      eprint={2302.04761},
      archivePrefix={arXiv},
      primaryClass={cs.CL},
      url={https://arxiv.org/abs/2302.04761}, 
}

@misc{schrader2025solverloop,
      title={A Solver-in-the-Loop Framework for Improving LLMs on Answer Set Programming for Logic Puzzle Solving}, 
      author={Timo Pierre Schrader and Lukas Lange and Tobias Kaminski and Simon Razniewski and Annemarie Friedrich},
      year={2025},
      eprint={2512.17093},
      archivePrefix={arXiv},
      primaryClass={cs.AI},
      url={https://arxiv.org/abs/2512.17093}, 
}

@misc{seely2025sudokubench,
      title={Sudoku-Bench: Evaluating creative reasoning with Sudoku variants}, 
      author={Jeffrey Seely and Yuki Imajuku and Tianyu Zhao and Edoardo Cetin and Llion Jones},
      year={2025},
      eprint={2505.16135},
      archivePrefix={arXiv},
      primaryClass={cs.AI},
      url={https://arxiv.org/abs/2505.16135}, 
}

@misc{shi2025judgeagent,
      title={JudgeAgent: Beyond Static Benchmarks for Knowledge-Driven and Dynamic LLM Evaluation}, 
      author={Zhichao Shi and Xuhui Jiang and Chengjin Xu and Cangli Yao and Shengjia Ma and Yinghan Shen and Zixuan Li and Jian Guo and Yuanzhuo Wang},
      year={2026},
      eprint={2509.02097},
      archivePrefix={arXiv},
      primaryClass={cs.CL},
      url={https://arxiv.org/abs/2509.02097}, 
}

@misc{srivastava2022bigbench,
      title={Beyond the Imitation Game: Quantifying and extrapolating the capabilities of language models}, 
      author={Aarohi Srivastava and Abhinav Rastogi and Abhishek Rao and Abu Awal Md Shoeb and Abubakar Abid and Adam Fisch and Adam R. Brown and Adam Santoro and Aditya Gupta and Adrià Garriga-Alonso and Agnieszka Kluska and Aitor Lewkowycz and Akshat Agarwal and Alethea Power and Alex Ray and Alex Warstadt and Alexander W. Kocurek and Ali Safaya and Ali Tazarv and Alice Xiang and Alicia Parrish and Allen Nie and Aman Hussain and Amanda Askell and Amanda Dsouza and Ambrose Slone and Ameet Rahane and Anantharaman S. Iyer and Anders Andreassen and Andrea Madotto and Andrea Santilli and Andreas Stuhlmüller and Andrew Dai and Andrew La and Andrew Lampinen and Andy Zou and Angela Jiang and Angelica Chen and Anh Vuong and Animesh Gupta and Anna Gottardi and Antonio Norelli and Anu Venkatesh and Arash Gholamidavoodi and Arfa Tabassum and Arul Menezes and Arun Kirubarajan and Asher Mullokandov and Ashish Sabharwal and Austin Herrick and Avia Efrat and Aykut Erdem and Ayla Karakaş and B. Ryan Roberts and Bao Sheng Loe and Barret Zoph and Bartłomiej Bojanowski and Batuhan Özyurt and Behnam Hedayatnia and Behnam Neyshabur and Benjamin Inden and Benno Stein and Berk Ekmekci and Bill Yuchen Lin and Blake Howald and Bryan Orinion and Cameron Diao and Cameron Dour and Catherine Stinson and Cedrick Argueta and César Ferri Ramírez and Chandan Singh and Charles Rathkopf and Chenlin Meng and Chitta Baral and Chiyu Wu and Chris Callison-Burch and Chris Waites and Christian Voigt and Christopher D. Manning and Christopher Potts and Cindy Ramirez and Clara E. Rivera and Clemencia Siro and Colin Raffel and Courtney Ashcraft and Cristina Garbacea and Damien Sileo and Dan Garrette and Dan Hendrycks and Dan Kilman and Dan Roth and Daniel Freeman and Daniel Khashabi and Daniel Levy and Daniel Moseguí González and Danielle Perszyk and Danny Hernandez and Danqi Chen and Daphne Ippolito and Dar Gilboa and David Dohan and David Drakard and David Jurgens and Debajyoti Datta and Deep Ganguli and Denis Emelin and Denis Kleyko and Deniz Yuret and Derek Chen and Derek Tam and Dieuwke Hupkes and Diganta Misra and Dilyar Buzan and Dimitri Coelho Mollo and Diyi Yang and Dong-Ho Lee and Dylan Schrader and Ekaterina Shutova and Ekin Dogus Cubuk and Elad Segal and Eleanor Hagerman and Elizabeth Barnes and Elizabeth Donoway and Ellie Pavlick and Emanuele Rodola and Emma Lam and Eric Chu and Eric Tang and Erkut Erdem and Ernie Chang and Ethan A. Chi and Ethan Dyer and Ethan Jerzak and Ethan Kim and Eunice Engefu Manyasi and Evgenii Zheltonozhskii and Fanyue Xia and Fatemeh Siar and Fernando Martínez-Plumed and Francesca Happé and Francois Chollet and Frieda Rong and Gaurav Mishra and Genta Indra Winata and Gerard de Melo and Germán Kruszewski and Giambattista Parascandolo and Giorgio Mariani and Gloria Wang and Gonzalo Jaimovitch-López and Gregor Betz and Guy Gur-Ari and Hana Galijasevic and Hannah Kim and Hannah Rashkin and Hannaneh Hajishirzi and Harsh Mehta and Hayden Bogar and Henry Shevlin and Hinrich Schütze and Hiromu Yakura and Hongming Zhang and Hugh Mee Wong and Ian Ng and Isaac Noble and Jaap Jumelet and Jack Geissinger and Jackson Kernion and Jacob Hilton and Jaehoon Lee and Jaime Fernández Fisac and James B. Simon and James Koppel and James Zheng and James Zou and Jan Kocoń and Jana Thompson and Janelle Wingfield and Jared Kaplan and Jarema Radom and Jascha Sohl-Dickstein and Jason Phang and Jason Wei and Jason Yosinski and Jekaterina Novikova and Jelle Bosscher and Jennifer Marsh and Jeremy Kim and Jeroen Taal and Jesse Engel and Jesujoba Alabi and Jiacheng Xu and Jiaming Song and Jillian Tang and Joan Waweru and John Burden and John Miller and John U. Balis and Jonathan Batchelder and Jonathan Berant and Jörg Frohberg and Jos Rozen and Jose Hernandez-Orallo and Joseph Boudeman and Joseph Guerr and Joseph Jones and Joshua B. Tenenbaum and Joshua S. Rule and Joyce Chua and Kamil Kanclerz and Karen Livescu and Karl Krauth and Karthik Gopalakrishnan and Katerina Ignatyeva and Katja Markert and Kaustubh D. Dhole and Kevin Gimpel and Kevin Omondi and Kory Mathewson and Kristen Chiafullo and Ksenia Shkaruta and Kumar Shridhar and Kyle McDonell and Kyle Richardson and Laria Reynolds and Leo Gao and Li Zhang and Liam Dugan and Lianhui Qin and Lidia Contreras-Ochando and Louis-Philippe Morency and Luca Moschella and Lucas Lam and Lucy Noble and Ludwig Schmidt and Luheng He and Luis Oliveros Colón and Luke Metz and Lütfi Kerem Şenel and Maarten Bosma and Maarten Sap and Maartje ter Hoeve and Maheen Farooqi and Manaal Faruqui and Mantas Mazeika and Marco Baturan and Marco Marelli and Marco Maru and Maria Jose Ramírez Quintana and Marie Tolkiehn and Mario Giulianelli and Martha Lewis and Martin Potthast and Matthew L. Leavitt and Matthias Hagen and Mátyás Schubert and Medina Orduna Baitemirova and Melody Arnaud and Melvin McElrath and Michael A. Yee and Michael Cohen and Michael Gu and Michael Ivanitskiy and Michael Starritt and Michael Strube and Michał Swędrowski and Michele Bevilacqua and Michihiro Yasunaga and Mihir Kale and Mike Cain and Mimee Xu and Mirac Suzgun and Mitch Walker and Mo Tiwari and Mohit Bansal and Moin Aminnaseri and Mor Geva and Mozhdeh Gheini and Mukund Varma T and Nanyun Peng and Nathan A. Chi and Nayeon Lee and Neta Gur-Ari Krakover and Nicholas Cameron and Nicholas Roberts and Nick Doiron and Nicole Martinez and Nikita Nangia and Niklas Deckers and Niklas Muennighoff and Nitish Shirish Keskar and Niveditha S. Iyer and Noah Constant and Noah Fiedel and Nuan Wen and Oliver Zhang and Omar Agha and Omar Elbaghdadi and Omer Levy and Owain Evans and Pablo Antonio Moreno Casares and Parth Doshi and Pascale Fung and Paul Pu Liang and Paul Vicol and Pegah Alipoormolabashi and Peiyuan Liao and Percy Liang and Peter Chang and Peter Eckersley and Phu Mon Htut and Pinyu Hwang and Piotr Miłkowski and Piyush Patil and Pouya Pezeshkpour and Priti Oli and Qiaozhu Mei and Qing Lyu and Qinlang Chen and Rabin Banjade and Rachel Etta Rudolph and Raefer Gabriel and Rahel Habacker and Ramon Risco and Raphaël Millière and Rhythm Garg and Richard Barnes and Rif A. Saurous and Riku Arakawa and Robbe Raymaekers and Robert Frank and Rohan Sikand and Roman Novak and Roman Sitelew and Ronan LeBras and Rosanne Liu and Rowan Jacobs and Rui Zhang and Ruslan Salakhutdinov and Ryan Chi and Ryan Lee and Ryan Stovall and Ryan Teehan and Rylan Yang and Sahib Singh and Saif M. Mohammad and Sajant Anand and Sam Dillavou and Sam Shleifer and Sam Wiseman and Samuel Gruetter and Samuel R. Bowman and Samuel S. Schoenholz and Sanghyun Han and Sanjeev Kwatra and Sarah A. Rous and Sarik Ghazarian and Sayan Ghosh and Sean Casey and Sebastian Bischoff and Sebastian Gehrmann and Sebastian Schuster and Sepideh Sadeghi and Shadi Hamdan and Sharon Zhou and Shashank Srivastava and Sherry Shi and Shikhar Singh and Shima Asaadi and Shixiang Shane Gu and Shubh Pachchigar and Shubham Toshniwal and Shyam Upadhyay and Shyamolima and Debnath and Siamak Shakeri and Simon Thormeyer and Simone Melzi and Siva Reddy and Sneha Priscilla Makini and Soo-Hwan Lee and Spencer Torene and Sriharsha Hatwar and Stanislas Dehaene and Stefan Divic and Stefano Ermon and Stella Biderman and Stephanie Lin and Stephen Prasad and Steven T. Piantadosi and Stuart M. Shieber and Summer Misherghi and Svetlana Kiritchenko and Swaroop Mishra and Tal Linzen and Tal Schuster and Tao Li and Tao Yu and Tariq Ali and Tatsu Hashimoto and Te-Lin Wu and Théo Desbordes and Theodore Rothschild and Thomas Phan and Tianle Wang and Tiberius Nkinyili and Timo Schick and Timofei Kornev and Titus Tunduny and Tobias Gerstenberg and Trenton Chang and Trishala Neeraj and Tushar Khot and Tyler Shultz and Uri Shaham and Vedant Misra and Vera Demberg and Victoria Nyamai and Vikas Raunak and Vinay Ramasesh and Vinay Uday Prabhu and Vishakh Padmakumar and Vivek Srikumar and William Fedus and William Saunders and William Zhang and Wout Vossen and Xiang Ren and Xiaoyu Tong and Xinran Zhao and Xinyi Wu and Xudong Shen and Yadollah Yaghoobzadeh and Yair Lakretz and Yangqiu Song and Yasaman Bahri and Yejin Choi and Yichi Yang and Yiding Hao and Yifu Chen and Yonatan Belinkov and Yu Hou and Yufang Hou and Yuntao Bai and Zachary Seid and Zhuoye Zhao and Zijian Wang and Zijie J. Wang and Zirui Wang and Ziyi Wu},
      year={2023},
      eprint={2206.04615},
      archivePrefix={arXiv},
      primaryClass={cs.CL},
      url={https://arxiv.org/abs/2206.04615}, 
}

@misc{statictodynamic2025,
      title={Recent Advances in Large Language Model Benchmarks against Data Contamination: From Static to Dynamic Evaluation}, 
      author={Simin Chen and Yiming Chen and Zexin Li and Yifan Jiang and Zhongwei Wan and Yixin He and Dezhi Ran and Tianle Gu and Haizhou Li and Tao Xie and Baishakhi Ray},
      year={2025},
      eprint={2502.17521},
      archivePrefix={arXiv},
      primaryClass={cs.LG},
      url={https://arxiv.org/abs/2502.17521}, 
}

@misc{suzgun2022bbh,
      title={Challenging BIG-Bench Tasks and Whether Chain-of-Thought Can Solve Them}, 
      author={Mirac Suzgun and Nathan Scales and Nathanael Schärli and Sebastian Gehrmann and Yi Tay and Hyung Won Chung and Aakanksha Chowdhery and Quoc V. Le and Ed H. Chi and Denny Zhou and Jason Wei},
      year={2022},
      eprint={2210.09261},
      archivePrefix={arXiv},
      primaryClass={cs.CL},
      url={https://arxiv.org/abs/2210.09261}, 
}

@misc{tang2022looppuzzles,
      title={A Framework for Loop and Path Puzzle Satisfiability NP-Hardness Results}, 
      author={Hadyn Tang},
      year={2022},
      eprint={2202.02046},
      archivePrefix={arXiv},
      primaryClass={cs.CC},
      url={https://arxiv.org/abs/2202.02046}, 
}

@misc{tyagi2024gridpuzzle,
      title={Step-by-Step Reasoning to Solve Grid Puzzles: Where do LLMs Falter?}, 
      author={Nemika Tyagi and Mihir Parmar and Mohith Kulkarni and Aswin RRV and Nisarg Patel and Mutsumi Nakamura and Arindam Mitra and Chitta Baral},
      year={2024},
      eprint={2407.14790},
      archivePrefix={arXiv},
      primaryClass={cs.CL},
      url={https://arxiv.org/abs/2407.14790}, 
}

@misc{wang2022selfconsistency,
      title={Self-Consistency Improves Chain of Thought Reasoning in Language Models}, 
      author={Xuezhi Wang and Jason Wei and Dale Schuurmans and Quoc Le and Ed Chi and Sharan Narang and Aakanksha Chowdhery and Denny Zhou},
      year={2023},
      eprint={2203.11171},
      archivePrefix={arXiv},
      primaryClass={cs.CL},
      url={https://arxiv.org/abs/2203.11171}, 
}

@misc{wang2023planandsolve,
      title={Plan-and-Solve Prompting: Improving Zero-Shot Chain-of-Thought Reasoning by Large Language Models}, 
      author={Lei Wang and Wanyu Xu and Yihuai Lan and Zhiqiang Hu and Yunshi Lan and Roy Ka-Wei Lee and Ee-Peng Lim},
      year={2023},
      eprint={2305.04091},
      archivePrefix={arXiv},
      primaryClass={cs.CL},
      url={https://arxiv.org/abs/2305.04091}, 
}

@misc{wei2022cot,
      title={Chain-of-Thought Prompting Elicits Reasoning in Large Language Models}, 
      author={Jason Wei and Xuezhi Wang and Dale Schuurmans and Maarten Bosma and Brian Ichter and Fei Xia and Ed Chi and Quoc Le and Denny Zhou},
      year={2023},
      eprint={2201.11903},
      archivePrefix={arXiv},
      primaryClass={cs.CL},
      url={https://arxiv.org/abs/2201.11903v6},
}

@misc{wei2025satbench,
      title={SATBench: Benchmarking LLMs' Logical Reasoning via Automated Puzzle Generation from SAT Formulas}, 
      author={Anjiang Wei and Yuheng Wu and Yingjia Wan and Tarun Suresh and Huanmi Tan and Zhanke Zhou and Sanmi Koyejo and Ke Wang and Alex Aiken},
      year={2025},
      eprint={2505.14615},
      archivePrefix={arXiv},
      primaryClass={cs.AI},
      url={https://arxiv.org/abs/2505.14615}, 
}

@misc{xu2024benchmarkleakage,
      title={Benchmarking Benchmark Leakage in Large Language Models}, 
      author={Ruijie Xu and Zengzhi Wang and Run-Ze Fan and Pengfei Liu},
      year={2024},
      eprint={2404.18824},
      archivePrefix={arXiv},
      primaryClass={cs.CL},
      url={https://arxiv.org/abs/2404.18824}, 
}

@misc{xu2024contaminationreview,
      title={Benchmark Data Contamination of Large Language Models: A Survey}, 
      author={Cheng Xu and Shuhao Guan and Derek Greene and M-Tahar Kechadi},
      year={2024},
      eprint={2406.04244},
      archivePrefix={arXiv},
      primaryClass={cs.CL},
      url={https://arxiv.org/abs/2406.04244}, 
}

@misc{xu2024theagentcompany,
      title={TheAgentCompany: Benchmarking LLM Agents on Consequential Real World Tasks}, 
      author={Frank F. Xu and Yufan Song and Boxuan Li and Yuxuan Tang and Kritanjali Jain and Mengxue Bao and Zora Z. Wang and Xuhui Zhou and Zhitong Guo and Murong Cao and Mingyang Yang and Hao Yang Lu and Amaad Martin and Zhe Su and Leander Maben and Raj Mehta and Wayne Chi and Lawrence Jang and Yiqing Xie and Shuyan Zhou and Graham Neubig},
      year={2025},
      eprint={2412.14161},
      archivePrefix={arXiv},
      primaryClass={cs.CL},
      url={https://arxiv.org/abs/2412.14161}, 
}

@misc{yao2023react,
      title={ReAct: Synergizing Reasoning and Acting in Language Models}, 
      author={Shunyu Yao and Jeffrey Zhao and Dian Yu and Nan Du and Izhak Shafran and Karthik Narasimhan and Yuan Cao},
      year={2023},
      eprint={2210.03629},
      archivePrefix={arXiv},
      primaryClass={cs.CL},
      url={https://arxiv.org/abs/2210.03629}, 
}

@misc{yao2024taubench,
      title={$\tau$-bench: A Benchmark for Tool-Agent-User Interaction in Real-World Domains}, 
      author={Shunyu Yao and Noah Shinn and Pedram Razavi and Karthik Narasimhan},
      year={2024},
      eprint={2406.12045},
      archivePrefix={arXiv},
      primaryClass={cs.AI},
      url={https://arxiv.org/abs/2406.12045}, 
}

@misc{yao2024tot,
      title={Tree of Thoughts: Deliberate Problem Solving with Large Language Models}, 
      author={Shunyu Yao and Dian Yu and Jeffrey Zhao and Izhak Shafran and Thomas L. Griffiths and Yuan Cao and Karthik Narasimhan},
      year={2023},
      eprint={2305.10601},
      archivePrefix={arXiv},
      primaryClass={cs.CL},
      url={https://arxiv.org/abs/2305.10601}, 
}

@techreport{yato2003complexity,
 author = {Yato, Takayuki and Seta, Takahiro},
 issue = {103(2002-AL-087)},
 month = {Nov},
 title = {Complexity and Completeness of Finding Another Solution and Its Application to Puzzles},
 url = {https://ipsj.ixsq.nii.ac.jp/records/31947},
 year = {2002}
}

@misc{zhang2025prm,
      title={The Lessons of Developing Process Reward Models in Mathematical Reasoning}, 
      author={Zhenru Zhang and Chujie Zheng and Yangzhen Wu and Beichen Zhang and Runji Lin and Bowen Yu and Dayiheng Liu and Jingren Zhou and Junyang Lin},
      year={2025},
      eprint={2501.07301},
      archivePrefix={arXiv},
      primaryClass={cs.CL},
      url={https://arxiv.org/abs/2501.07301}, 
}

@misc{zhang2025puzzlebench,
      title={PuzzleBench: A Fully Dynamic Evaluation Framework for Large Multimodal Models on Puzzle Solving}, 
      author={Zeyu Zhang and Zijian Chen and Zicheng Zhang and Yuze Sun and Yuan Tian and Ziheng Jia and Chunyi Li and Xiaohong Liu and Xiongkuo Min and Guangtao Zhai},
      year={2025},
      eprint={2504.10885},
      archivePrefix={arXiv},
      primaryClass={cs.CV},
      url={https://arxiv.org/abs/2504.10885}, 
}

@misc{zheng2025prmsurvey,
      title={A Survey of Process Reward Models: From Outcome Signals to Process Supervisions for Large Language Models}, 
      author={Congming Zheng and Jiachen Zhu and Zhuoying Ou and Yuxiang Chen and Kangning Zhang and Rong Shan and Zeyu Zheng and Mengyue Yang and Jianghao Lin and Yong Yu and Weinan Zhang},
      year={2025},
      eprint={2510.08049},
      archivePrefix={arXiv},
      primaryClass={cs.CL},
      url={https://arxiv.org/abs/2510.08049}, 
}

@misc{zhou2022leasttomost,
      title={Least-to-Most Prompting Enables Complex Reasoning in Large Language Models}, 
      author={Denny Zhou and Nathanael Schärli and Le Hou and Jason Wei and Nathan Scales and Xuezhi Wang and Dale Schuurmans and Claire Cui and Olivier Bousquet and Quoc Le and Ed Chi},
      year={2023},
      eprint={2205.10625},
      archivePrefix={arXiv},
      primaryClass={cs.AI},
      url={https://arxiv.org/abs/2205.10625v3},
}


\appendix

\section*{Appendix A: Puzzle Variety Descriptions}
\label{app:puzzle_types}

Brief descriptions of the \numpuzzletypes{} varieties in the evaluation set. The full dataset contains \valFullVarieties{} varieties.

\paragraph{Country}
Draw a loop through orthogonally adjacent cells that visits every outlined country exactly once. Numbers indicate how many cells the loop passes through in that country. The loop cannot branch or cross itself.

\paragraph{Dbchoco}
Divide the grid into regions. Each region contains one white and one grey contiguous area of identical size and shape (allowing rotation/reflection). Numbers indicate the size of the area they're in.

\paragraph{Firefly}
Draw paths from every firefly (starting at black dots) to form one connected network. Paths cannot branch, cross, or connect directly between two black dots. Numbers indicate how many times the path turns.

\paragraph{Heyawake}
Shade some cells in a grid divided into rooms. Shaded cells cannot be adjacent. Numbers indicate shaded cells in that room. No horizontal/vertical line of unshaded cells can pass through 2+ room borders.

\paragraph{Hitori}
Shade some cells so that no row or column contains duplicate unshaded numbers. Shaded cells cannot be adjacent. All unshaded cells must form one orthogonally connected area.

\paragraph{Kurodoko}
Shade some cells (not numbers). Shaded cells cannot be adjacent. Numbers indicate how many unshaded cells can be seen in a straight line horizontally and vertically, including itself. All unshaded cells must connect.

\paragraph{Lightup}
Place lights in empty cells to illuminate all non-black cells. Lights illuminate in straight lines until blocked by black cells. Lights cannot illuminate each other. Numbers on black cells indicate adjacent lights.

\paragraph{Lits}
Place one tetromino (4-cell block) in every outlined region. No 2x2 square can be fully covered. Identical tetrominoes (including rotations/reflections) cannot share an edge. All tetrominoes must connect.

\paragraph{Mashu}
Draw a loop through every circle. On black circles: the loop must turn and go straight before/after. On white circles: the loop must go straight but turn in at least one adjacent cell.

\paragraph{Norinori}
Shade cells so that each shaded cell is adjacent to exactly one other shaded cell (forming dominoes). Each outlined region contains exactly 2 shaded cells.

\paragraph{Nurikabe}
Shade cells to form regions of unshaded cells. Each region contains exactly one number indicating its size. Numbers cannot be shaded. Shaded cells cannot form 2x2 squares and must all connect.

\paragraph{Nurimaze}
Shade some tiles to form a maze. Tiles are fully shaded or unshaded. Clued tiles cannot be shaded. No 2x2 square can be all shaded or all unshaded. Unshaded cells form a path from S to G passing through all circles.

\paragraph{Nurimisaki}
Shade cells with no 2x2 squares of same color. Circles mark cells that are unshaded and adjacent to exactly one other unshaded cell. Numbers indicate visible unshaded cells in a straight line including itself.

\paragraph{Sashigane}
Divide the grid into L-shaped regions (width one cell). Circles must be at the corner of an L. Arrows must be at the ends, pointing toward the corner. Numbers indicate total cells in that L-shape.

\paragraph{Shakashaka}
Place right triangles (half-cells) so that every unshaded area forms a rectangle (upright or 45\textdegree{} rotated). Numbers indicate how many adjacent cells contain triangles.

\paragraph{Shikaku}
Divide the grid into rectangles. Each rectangle contains exactly one number indicating its size in cells.

\paragraph{Slither}
Draw a single loop along cell edges. Numbers indicate how many of that cell's four edges are part of the loop. The loop cannot branch or cross.

\paragraph{Sudoku}
Place numbers 1-N in each cell (N = grid width). Each row, column, and outlined block contains exactly one of each number.

\paragraph{Tapa}
Shade cells (not numbers). Numbers indicate lengths of consecutive shaded cell blocks in the 8 surrounding cells. Shaded cells cannot form 2x2 squares and must all connect.

\paragraph{Yajilin}
Shade some cells and draw a loop through the rest. Shaded cells cannot be adjacent. Number clues cannot be shaded and indicate how many shaded cells lie in the arrow's direction.

\section*{Appendix B: Prompt Templates}
\label{app:prompts}

\subsection*{B.1 DirectAsk (Single-Shot) Strategy}

\textbf{System Prompt:}
\begin{lstlisting}
Solve the puzzle!!

Answer with a list of moves you would like to make that
solve the puzzle as json in a markdown json code block

```json
["mouse,left,1,1", "mouse,right,3,1", ... ]
```
\end{lstlisting}

\textbf{User Prompt Template:}
\begin{lstlisting}
Puzzle Type: {puzzle_type}
Puzzle Rules:
```
{rules_text}
```

Here is an example of inputs / a solved puzzle
  (lots of context for you)
```
{example_of_inputs}
```
Here's some more:
{example_move_context}
Note specifically how the coordinate systems work
  (for the puzzle vs. the inputs).
For the puzzle you are working on, ensure you fully
  understand from the example of input above, that
  the move is exactly where you expect.
==== ==== ==== ==== ====
Here is the puzzle you are to solve:
{puzzle_state}

==== ==== ==== ==== ====
Please now solve it.
\end{lstlisting}

\subsection*{B.2 BasicAgentic (Multi-Turn) Strategy}

\textbf{System Prompt:}
\begin{lstlisting}
Solve the puzzle!!
It is known to be solvable, and you can figure it out.
This is a logic deduction benchmark.
You are graded on both how many steps you take (number of
tool calls), how many moves it takes (puzzle moves), and
ultimately if you can solve the puzzle. This puzzle is
solvable. If you need to, you may reset the puzzle and
keep trying.
\end{lstlisting}

\textbf{User Prompt:} Same template as DirectAsk (Section B.1).

\textbf{Available Tools:}
\begin{itemize}
\item \texttt{make\_move(movestring)}: ``Make a move, shows the board after the move is applied''
\item \texttt{make\_multi\_move(movelist)}: ``Make a series of moves, shows the board after the move is applied''
\item \texttt{check\_board\_for\_completeness()}: ``Check the current state of the board against the rules of the puzzle, see if its complete or if errors exist''
\item \texttt{render\_board\_as\_svg()}: ``Shows the full detail SVG of the board (useful if you want more information / are worried about your view into errors)''
\item \texttt{get\_rules()}: ``Gets all the rules for the puzzle''
\item \texttt{reset\_puzzle()}: ``Fully reset the puzzle (erase all moves, go back to a blank slate). Use this instead of giving up if you want another attempt.''
\item \texttt{give\_up()}: Forfeit the attempt
\end{itemize}

\textbf{Output Validation Loop:} If the model produces a text response before the puzzle is complete, the agent framework automatically retries with the message: ``Not done yet, keep going!! Puzzle isn't complete.'' This continues until the puzzle is solved, the model calls \texttt{give\_up()}, or the maximum move limit (5000) is reached.

\section*{Appendix C: Full Model Results}
\label{app:full_results}

Complete results for all \nummodels{} evaluated models across both strategies.

\begin{table}[ht!]
\centering
\caption{Full results by model and strategy. Succ = correct solves, Fail = wrong answers, Err = infrastructure failures (timeouts, API errors), \$/Att = cost per attempt. Strategies are evaluated on different puzzle sets and rates are \textbf{not} comparable across columns.}
\label{tab:full_results}
\scriptsize
\setlength{\tabcolsep}{2.5pt}
\begin{tabular}{lrrrrrrrrrr}
\toprule
 & \multicolumn{5}{c}{Direct Ask (\valGoldenPuzzles{} puzzles)} & \multicolumn{5}{c}{Agentic (\valGoldenThirtyPuzzles{}/\valGoldenSixtyPuzzles{} puzzles)} \\
\cmidrule(lr){2-6} \cmidrule(lr){7-11}
Model & Succ & Fail & Err & Rate & \$/Att & Succ & Fail & Err & Rate & \$/Att \\
\midrule
gpt-5.2@xhigh & 81 & 87 & 132 & 27.0\% & \$3.76 & 47 & 8 & 29 & \textbf{56.0\%} & \$9.74 \\
claude-opus-4-6-1m & 0 & 300 & 0 & 0.0\% & \$0.08 & 11 & 19 & 0 & \textbf{36.7\%} & \$14.11 \\
gpt-5.2@high & 62 & 235 & 3 & 20.7\% & \$1.07 & 11 & 18 & 1 & \textbf{36.7\%} & \$7.30 \\
claude-opus-4-6@thinking & 82 & 207 & 11 & 27.3\% & \$3.18 & 28 & 23 & 33 & \textbf{33.3\%} & \$6.24 \\
gemini-3.1-pro & 60 & 240 & 0 & 20.0\% & \$0.41 & 28 & 47 & 9 & \textbf{33.3\%} & \$14.36 \\
claude-opus-4-6 & 1 & 299 & 0 & 0.3\% & \$0.05 & 9 & 1 & 20 & \textbf{30.0\%} & \$10.89 \\
claude-sonnet-4-6@thinking & 31 & 263 & 6 & 10.3\% & \$0.56 & 8 & 0 & 22 & \textbf{26.7\%} & \$3.94 \\
gpt-5.2-pro & 29 & 137 & 134 & 9.7\% & \$2.70 & 8 & 12 & 10 & \textbf{26.7\%} & \$41.52 \\
claude-opus-4-6@max & 1 & 299 & 0 & 0.3\% & \$0.04 & 7 & 5 & 18 & \textbf{23.3\%} & \$10.87 \\
claude-sonnet-4-6-1m & 1 & 299 & 0 & 0.3\% & \$0.05 & 7 & 3 & 20 & \textbf{23.3\%} & \$169.27 \\
gpt-5.2@medium & 28 & 272 & 0 & 9.3\% & \$0.47 & 7 & 23 & 0 & \textbf{23.3\%} & \$2.81 \\
claude-sonnet-4-6 & 1 & 299 & 0 & 0.3\% & \$0.03 & 5 & 0 & 25 & \textbf{16.7\%} & \$8.91 \\
gemini-3-pro@high & 10 & 290 & 0 & 3.3\% & \$0.33 & 5 & 25 & 0 & \textbf{16.7\%} & \$10.72 \\
gemini-3-pro & 13 & 287 & 0 & 4.3\% & \$0.26 & 4 & 26 & 0 & \textbf{13.3\%} & \$4.29 \\
gemini-3-pro@minimal & 12 & 287 & 1 & 4.0\% & \$0.27 & 3 & 27 & 0 & \textbf{10.0\%} & \$6.62 \\
gpt-5.2@low & 7 & 293 & 0 & 2.3\% & \$0.07 & 3 & 27 & 0 & \textbf{10.0\%} & \$0.60 \\
gpt-5.1@medium & 23 & 277 & 0 & \textbf{7.7\%} & \$0.22 & 2 & 28 & 0 & 6.7\% & \$0.81 \\
claude-opus-4-5@thinking & 18 & 272 & 10 & 6.0\% & \$0.78 & 2 & 26 & 2 & \textbf{6.7\%} & \$4.71 \\
gemini-3-flash@high & 9 & 291 & 0 & 3.0\% & \$0.09 & 2 & 28 & 0 & \textbf{6.7\%} & \$0.90 \\
gemini-3-flash@minimal & 14 & 285 & 1 & 4.7\% & \$0.09 & 2 & 27 & 1 & \textbf{6.7\%} & \$0.97 \\
gpt-5@medium & 18 & 282 & 0 & \textbf{6.0\%} & \$0.18 & 1 & 29 & 0 & 3.3\% & \$11.21 \\
kimi-k2.5 & 18 & 282 & 0 & \textbf{6.0\%} & \$0.23 & 1 & 0 & 29 & 3.3\% & \$1.60 \\
grok-4-1-fast & 17 & 283 & 0 & \textbf{5.7\%} & \$0.001 & 1 & 19 & 10 & 3.3\% & \$0.41 \\
grok-4-1-fast-reasoning & 16 & 283 & 1 & \textbf{5.3\%} & \$0.001 & 0 & 18 & 12 & 0.0\% & \$0.46 \\
claude-opus-4-5-high & 1 & 299 & 0 & 0.3\% & \$0.04 & 1 & 25 & 4 & \textbf{3.3\%} & \$8.82 \\
claude-sonnet-4-5 & 0 & 300 & 0 & 0.0\% & \$0.03 & 1 & 10 & 19 & \textbf{3.3\%} & \$12.20 \\
minimax-m2.5 & 2 & 237 & 61 & 0.7\% & \$0.07 & 1 & 17 & 12 & \textbf{3.3\%} & \$1.95 \\
o3 & 9 & 291 & 0 & 3.0\% & \$0.19 & 1 & 29 & 0 & \textbf{3.3\%} & \$2.45 \\
claude-sonnet-4-5@thinking & 7 & 293 & 0 & \textbf{2.3\%} & \$0.40 & 0 & 16 & 14 & 0.0\% & \$7.49 \\
deepseek-v3.2 & 6 & 289 & 5 & \textbf{2.0\%} & \$0.04 & 0 & 27 & 3 & 0.0\% & \$1.58 \\
deepseek-v3.2-speciale & 6 & 233 & 61 & \textbf{2.0\%} & \$0.10 & -- & -- & -- & -- & -- \\
kimi-k2-thinking & 4 & 296 & 0 & \textbf{1.3\%} & \$0.20 & 0 & 23 & 7 & 0.0\% & \$1.00 \\
glm-5 & 2 & 298 & 0 & \textbf{0.7\%} & \$0.10 & 0 & 28 & 2 & 0.0\% & \$8.46 \\
o1 & 2 & 298 & 0 & \textbf{0.7\%} & \$0.51 & 0 & 30 & 0 & 0.0\% & \$4.03 \\
qwen3.5-397b-a17b & 2 & 296 & 2 & \textbf{0.7\%} & \$0.08 & 0 & 0 & 30 & 0.0\% & \$0.000 \\
gemini-2.5-pro & 1 & 299 & 0 & \textbf{0.3\%} & \$0.28 & 0 & 30 & 0 & 0.0\% & \$1.96 \\
glm-4.7 & 1 & 299 & 0 & \textbf{0.3\%} & \$0.08 & 0 & 29 & 1 & 0.0\% & \$1.73 \\
gpt-5.2 & 1 & 299 & 0 & \textbf{0.3\%} & \$0.007 & 0 & 30 & 0 & 0.0\% & \$0.61 \\
gpt-oss-120b & 1 & 299 & 0 & \textbf{0.3\%} & \$0.002 & -- & -- & -- & -- & -- \\
grok-code-fast-1 & 1 & 297 & 2 & \textbf{0.3\%} & \$0.004 & 0 & 17 & 13 & 0.0\% & \$0.23 \\
mimo-v2-flash & 1 & 296 & 3 & \textbf{0.3\%} & \$0.01 & 0 & 22 & 8 & 0.0\% & \$0.94 \\
minimax-m2.1 & 1 & 299 & 0 & \textbf{0.3\%} & \$0.08 & 0 & 24 & 6 & 0.0\% & \$1.67 \\
qwen3-235b-a22b-thinking-2507 & 1 & 299 & 0 & \textbf{0.3\%} & \$0.02 & 0 & 29 & 1 & 0.0\% & \$0.71 \\
qwen3-next-80b-a3b-thinking & 1 & 298 & 1 & \textbf{0.3\%} & \$0.009 & 0 & 8 & 22 & 0.0\% & \$2.62 \\
qwen3-vl-235b-a22b-thinking & 1 & 294 & 5 & \textbf{0.3\%} & \$0.06 & -- & -- & -- & -- & -- \\
devstral-2512 & 0 & 300 & 0 & 0.0\% & \$0.001 & 0 & 25 & 5 & 0.0\% & \$0.20 \\
gpt-3.5-turbo & 0 & 300 & 0 & 0.0\% & \$0.002 & 0 & 0 & 30 & 0.0\% & \$0.000 \\
gpt-4.1 & 0 & 300 & 0 & 0.0\% & \$0.009 & 0 & 19 & 11 & 0.0\% & \$214.97 \\
gpt-4o & 0 & 300 & 0 & 0.0\% & \$0.01 & 0 & 22 & 8 & 0.0\% & \$55.57 \\
mistral-large-2512 & 0 & 300 & 0 & 0.0\% & \$0.003 & 0 & 26 & 4 & 0.0\% & \$38.65 \\
qwen3-coder & 0 & 300 & 0 & 0.0\% & \$0.001 & 0 & 30 & 0 & 0.0\% & \$0.68 \\
\bottomrule

\end{tabular}
\end{table}

\section*{Appendix D: Puzzle Variety Gallery}
\label{app:gallery}

\begin{figure}[t]
\centering
\includegraphics[width=\textwidth]{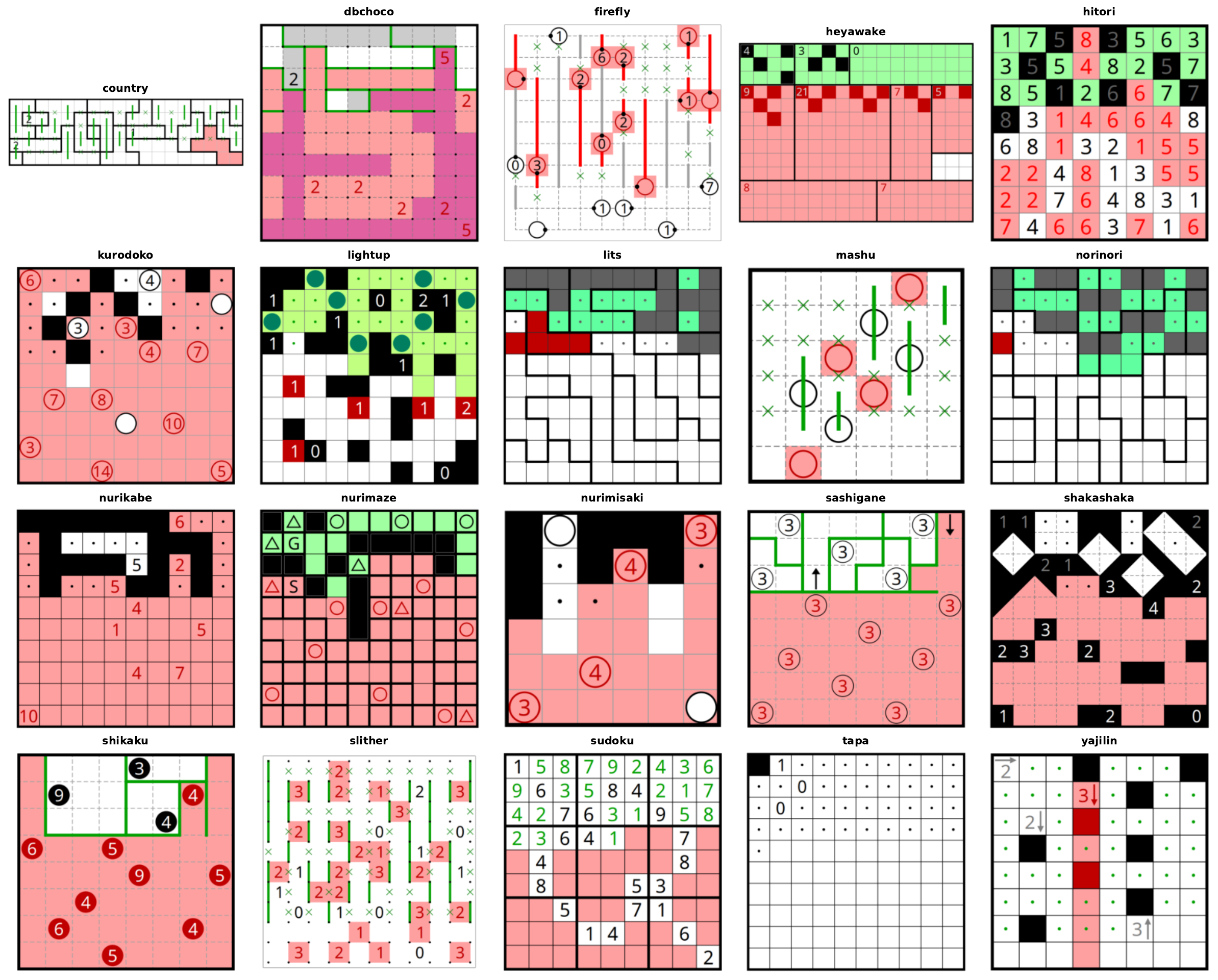}
\caption{All \numpuzzletypes{} varieties in the benchmark, shown mid-solve with constraint violations highlighted in red.}
\label{fig:puzzle_gallery}
\end{figure}

Figure~\ref{fig:puzzle_gallery} shows one puzzle from each of the \numpuzzletypes{} benchmark varieties, each partially solved. Red highlighting indicates cells where variety-specific constraints are currently violated.

\section*{Appendix E: Full Model Success Over Time}
\label{app:full_timeline}

\begin{figure}[t]
\centering
\includegraphics[width=0.95\textwidth]{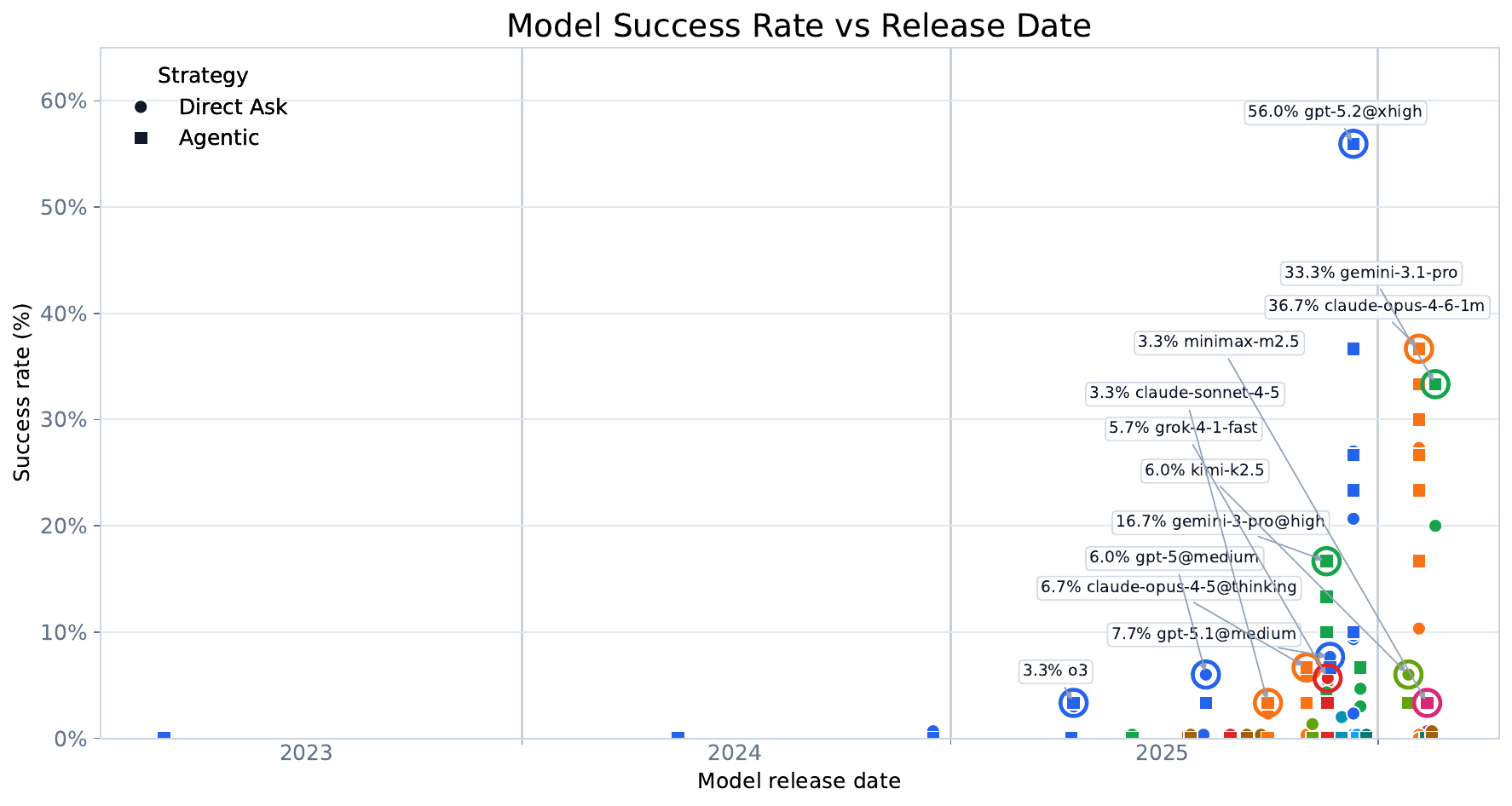}
\caption{Full model success rates over time with frontier model release dates annotated, including all evaluated models. Models released before late 2024 show 0\% or near-0\% solve rates, with capability emerging only in late 2025.}
\label{fig:full_timeline}
\end{figure}

Figure~\ref{fig:full_timeline} extends the timeline from the main text (Figure~\ref{fig:model_timeline}) to include all evaluated models, spanning release dates from 2022 through early 2026. The extended view reveals that \textbf{pencil puzzle solving capability is entirely a recent phenomenon}: models released before late 2024---including GPT-3.5-turbo, GPT-4o, GPT-4.1, and o1---achieve 0\% or near-0\% solve rates. The first non-trivial performance appeared with o3 (3.0\%) in early 2025, followed by an explosion in capability starting around November 2025 with the GPT-5 family and reasoning-enabled models. This back-testing confirms that our benchmark measures a genuinely new capability frontier, not one that was latent in earlier model generations.

\end{document}